\documentclass[]{youtu} 

\usepackage{mathpazo}
\usepackage{amssymb}
\usepackage{graphicx}
\usepackage{natbib}
\usepackage{multirow}
\usepackage{makecell}
\usepackage{array}
\usepackage{xcolor} 
\usepackage{colortbl} 
\usepackage{booktabs} 
\usepackage{CJKutf8} 
\usepackage{float} 
\title{Youtu-Parsing: Perception, Structuring and Recognition via High-Parallelism Decoding}

\author{Youtu-Parsing Team$^*$}

\renewcommand{\thefootnote}{*}
\footnotetext{Full author list in contributions.}
\renewcommand{\thefootnote}{\arabic{footnote}}

\sourcecode{https://github.com/TencentCloudADP/youtu-parsing}
\model{https://huggingface.co/collections/tencent/youtu}
\demo{https://huggingface.co/spaces/Tencent/Youtu-Parsing}

\begin{document}
\begin{CJK}{UTF8}{gbsn}

\abstract{

This paper presents Youtu-Parsing, an efficient and versatile document parsing model designed for high-performance content extraction. The architecture employs a native Vision Transformer (ViT) featuring a dynamic-resolution visual encoder to extract shared document features, coupled with a prompt-guided Youtu-LLM-2B language model for layout analysis and region-prompted decoding. Leveraging this decoupled and feature-reusable framework, we introduce a high-parallelism decoding strategy comprising two core components: token parallelism and query parallelism. 
The token parallelism strategy concurrently generates up to 64 candidate tokens per inference step, which are subsequently validated through a verification mechanism. This approach yields a $5\text{--}11\times$ speedup over traditional autoregressive decoding and is particularly well-suited for highly structured scenarios, such as table recognition. To further exploit the advantages of region-prompted decoding, the query parallelism strategy enables simultaneous content prediction for multiple bounding boxes (up to five), providing an additional $2\times$ acceleration while maintaining output quality equivalent to standard decoding. 
Youtu-Parsing encompasses a diverse range of document elements, including text, formulas, tables, charts, seals, and hierarchical structures. Furthermore, the model exhibits strong robustness when handling rare characters, multilingual text, and handwritten content. Extensive evaluations demonstrate that Youtu-Parsing achieves state-of-the-art (SOTA) performance on both the OmniDocBench and olmOCR-bench benchmarks. Overall, Youtu-Parsing demonstrates significant experimental value and practical utility for large-scale document intelligence applications.

}

\maketitle

\vspace{-.1em}


\begin{figure}[htbp]
  \centering
  \includegraphics[width=1.0\textwidth]{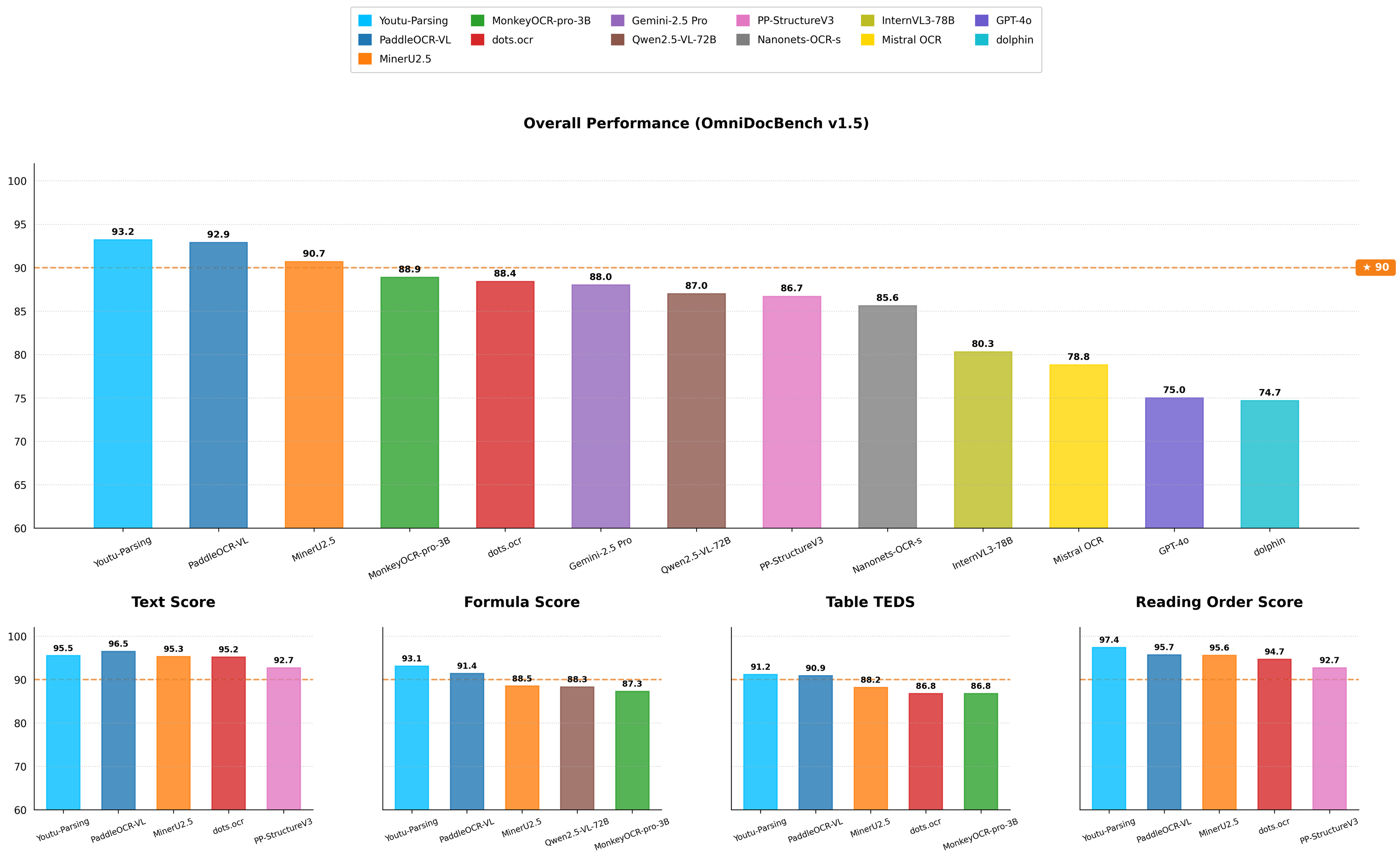}
  \caption{Performance of Youtu-Parsing on OmniDocBench v1.5. Youtu-Parsing surpasses both general-purpose vision-language models and specialized domain models, setting new benchmarks in text recognition, formula recognition, table recognition, and reading order prediction across multiple evaluation tasks.}
  \label{fig:score}
\end{figure}

\begin{figure}[htbp]
  \centering
  \includegraphics[width=1.0\textwidth]{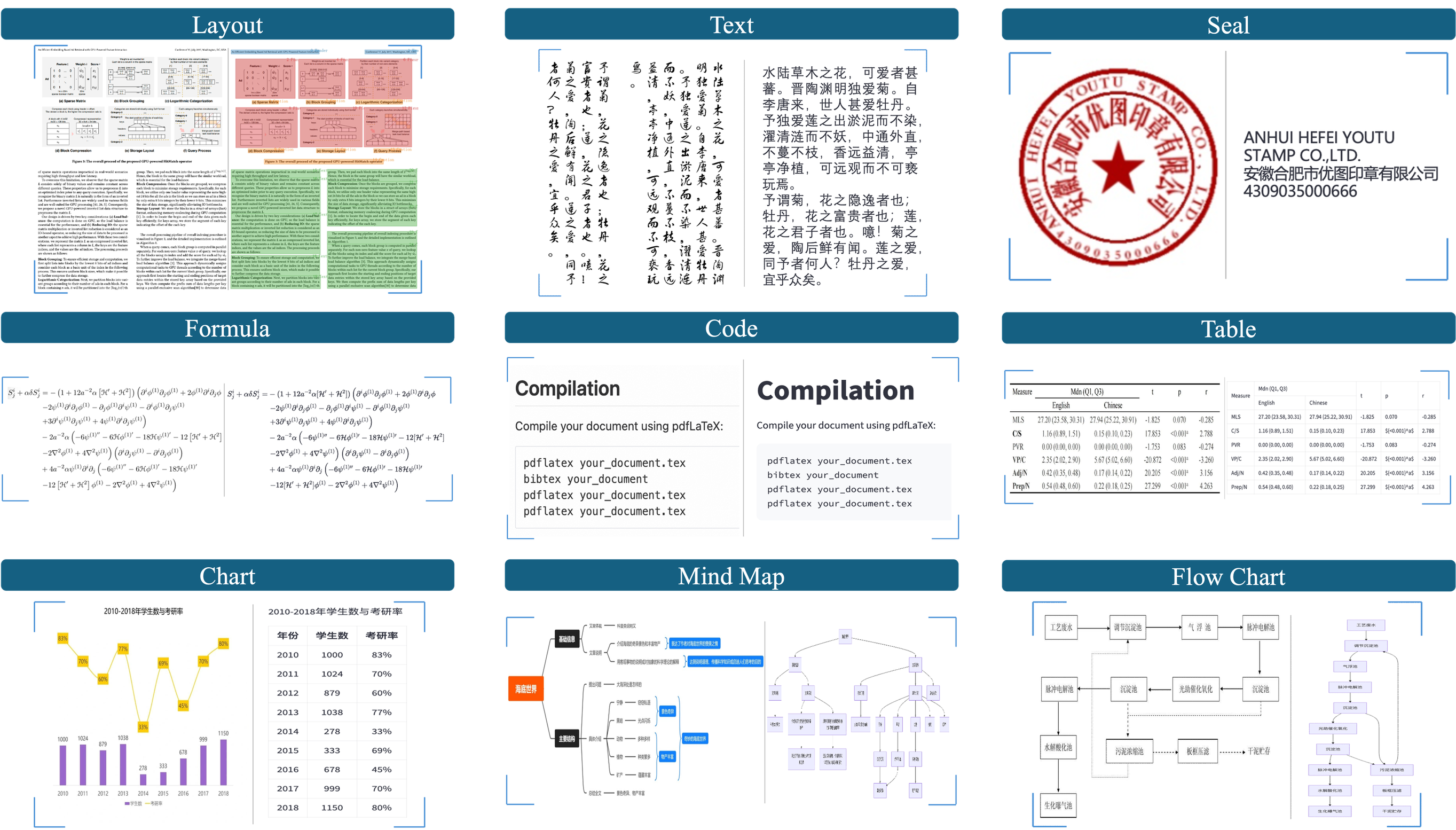}
  \caption{Qualitative parsing samples of Youtu-Parsing across diverse document elements, including layout, text, seal, formula, code, table, chart, mind map, and flow chart. Each panel shows a representative input alongside its structured parsing result.}
  \label{fig:sample}
\end{figure}

\section{Introduction}

Driven by the exponential expansion of digital information, documents have emerged as indispensable repositories for knowledge storage and transmission across diverse domains. However, the escalating complexity and volume of modern documents pose substantial challenges for effective information extraction, necessitating the advancement of sophisticated document parsing \citep{zhang2024document, feng2025dolphin, li2025monkeyocr, wang2024mineru} techniques. The primary objectives of document parsing are manifold: precisely identifying and segmenting structural components—such as text blocks, columns, mathematical formulas, tables, and figures; establishing a logical reading order to preserve semantic coherence; and detecting auxiliary elements including footnotes and captions. Successfully achieving these goals is pivotal for facilitating efficient information retrieval and empowering downstream applications, such as content summarization, knowledge graph construction, and question answering.

\begin{table}[htbp]
\setcitestyle{numbers, square} 
\caption{Supported elements for parsing.}
\label{tab:supported-elements}
\centering
\small
\begin{tabular}{lccccccc}
\toprule
\textbf{Models}  & \textbf{\makecell{Text}} & \textbf{\makecell{Formula}} & \textbf{\makecell{Table}} & \textbf{\makecell{Chart}} &  \textbf{\makecell{Seal}} & \textbf{\makecell{Hierarchical \\ Structure}} \\
\midrule
olmOCR-7B \citep{poznanski2025olmocr} & \checkmark & \checkmark & \checkmark & & &  \\
Dolphin \citep{feng2025dolphin}  & \checkmark & \checkmark & \checkmark & & &  \\
MonkeyOCR-pro-3B \citep{li2025monkeyocr} & \checkmark & \checkmark & \checkmark & & &  \\
dots.ocr \citep{li2025dots} & \checkmark & \checkmark & \checkmark & & &  \\
MinerU2.5 \citep{niu2025mineru2} & \checkmark & \checkmark & \checkmark & & &  \\
PaddleOCR-VL \citep{cui2025paddleocrvl} & \checkmark & \checkmark & \checkmark & \checkmark & &  \\
Youtu-Parsing   & \checkmark & \checkmark & \checkmark & \checkmark & \checkmark & \checkmark \\
\bottomrule
\end{tabular}
\end{table}

The rapid evolution of Large Language Models (LLMs) \citep{brown2020language, touvron2023llama, achiam2023gpt, team2023gemini, jiang2023mistral} and Vision-Language Models (VLMs) \citep{radford2021learning, liu2023visual, zhu2023minigpt, dai2023instructblip, tschannen2025siglip} has catalyzed significant progress in document parsing. Nevertheless, the field continues to grapple with the challenge of balancing high-fidelity recognition with the stringent efficiency requirements of real-world applications. Modern documents encompass a wide array of formats and content types—such as academic publications, legal filings, and presentation slides—that integrate heterogeneous elements including text, tables, formulas, and seals. Accurately parsing these diverse data types requires sophisticated spatial-semantic reasoning, yet practical deployment further demands low inference latency to ensure scalability. Consequently, achieving robust generalization across varied document categories while maintaining high computational efficiency remains a critical bottleneck in the development of universal parsing frameworks.

Current methodologies are generally classified into two paradigms: modular pipeline-based systems \citep{cui2025paddleocr,wang2024mineru,livathinos2025docling, niu2025mineru2} and end-to-end multimodal models \citep{li2025dots,poznanski2025olmocr,wei2025deepseek}. While pipeline-based methods offer interpretability and modular flexibility, they are often hampered by cumbersome workflows, intensive manual tuning, and the risk of cascading errors across successive stages. Conversely, end-to-end models facilitate joint optimization but frequently encounter difficulties in processing long-form documents, mitigating hallucinations (i.e., the generation of erroneous or fabricated content), and maintaining computational efficiency. Overcoming these limitations is imperative for developing next-generation document parsing technologies that are accurate, efficient, and robust across diverse document corpora.

To address these challenges, we introduce \textbf{Youtu-Parsing}, a novel, high-efficiency, and decoupled document parsing framework specifically engineered for complex real-world scenarios. The framework decomposes the document parsing task into three synergistic stages: \textbf{shared visual feature extraction}, \textbf{layout analysis}, and \textbf{region-prompted decoding}. The shared visual feature extraction module leverages NaViT \citep{tschannen2025siglip} to generate a shared visual feature map, which serves as a unified representation for all subsequent parsing operations. Building upon this, the layout analysis module executes rapid structural parsing to precisely identify both the bounding box coordinates and semantic categories of document elements. Finally, the region-based content querying module extracts textual information for each detected element. By employing category-specific prompts, this module effectively mitigates stylistic interference arising from the heterogeneous output requirements of different element types.
This decoupled yet integrated architecture circumvents the error accumulation typical of traditional pipelines while facilitating modular training and targeted optimization for individual sub-tasks. Furthermore, by querying the content of layout elements independently, the framework inherently supports highly parallelized decoding, thereby substantially enhancing the overall inference throughput and efficiency of the document parsing process.

Document parsing tasks, particularly those centered on Optical Character Recognition (OCR), are characterized by a high degree of output determinism. Unlike open-ended natural language generation, the tokens in document parsing are rigorously grounded in visual cues, and spatial dependencies between distant tokens are relatively sparse. Drawing inspiration from existing literature on efficient sequence generation \citep{ran2020learning, chang2022maskgit, cai2024medusa}, we leverage these properties to propose a dual-track parallelization paradigm: \textbf{Token Parallelism} and \textbf{Query Parallelism}.

\paragraph{Token Parallelism}
Conventional autoregressive decoding generates tokens sequentially, which constitutes a significant throughput bottleneck in large-scale document digitization. To circumvent this, we introduce a lossless token-level parallel decoding scheme consisting of two core operations:
\begin{enumerate}
    \item \textbf{Candidate Generation:} In each inference iteration, the model simultaneously predicts a block of $N$ candidate tokens (up to 64) by extending the input prefix with specialized mask tokens.
    \item \textbf{Decoding Verification:} The generated candidates are validated through a verification mechanism to ensure the output is identical to that of standard autoregressive decoding. This guarantees zero degradation in recognition accuracy while maintaining mathematical equivalence to the baseline.
\end{enumerate}
The decoding process iterates until the End-of-Sequence (EOS) token is generated and verified, signaling the completion of the sequence.

To facilitate this capability, we employ a \textbf{Hybrid Masked Training (HMT)} strategy. During the fine-tuning phase, 80\% of the training samples are augmented with masks of random positions and lengths, incentivizing the model to capture multi-token look-ahead dependencies. The remaining 20\% of samples remain unmasked to preserve the integrity of standard autoregressive performance. This strategy yields a $5\text{--}11\times$ empirical speedup, consistent with the theoretical acceleration $S \approx k/2$, where $k$ denotes the average number of accepted tokens per iteration.

\paragraph{Query Parallelism}
Existing Vision-Language Models (VLMs) typically process bounding boxes sequentially, leading to significant computational redundancy. Leveraging our decoupled architecture where visual features are shared, Youtu-Parsing inherently supports \textit{Query Parallelism}. 

Specifically, we feed multiple bounding boxes (up to five) into the LLM module concurrently. The model generates a concatenated output sequence, which is subsequently partitioned and mapped to the corresponding layout elements. This approach offers two primary advantages:
\begin{itemize}
    \item \textbf{Concurrent Extraction:} It enables the simultaneous recognition of multiple independent regions, such as disparate formulas or text blocks.
    \item \textbf{Tail-End Optimization:} It effectively utilizes the residual computational capacity during the token-parallel decoding process.
\end{itemize}

By querying the maximum capacity of five bounding boxes simultaneously, Youtu-Parsing achieves an approximately $2\times$ throughput improvement compared to sequential box processing, significantly enhancing its efficiency for large-scale document analysis.

Youtu-Parsing supports a comprehensive array of document components, including text, formulas, tables, charts, seals, and hierarchical structures. As summarized in Table \ref{tab:supported-elements}, this integrated capability enables the framework to interpret diverse real-world documents—such as academic publications, legal filings, and presentation slides—with high precision. By providing unified support for these heterogeneous elements, Youtu-Parsing facilitates a holistic approach to document digitization, ensuring high-fidelity results across a wide spectrum of complex scenarios.

Youtu-Parsing addresses the critical challenges in document processing through the following key contributions:

\begin{itemize}
    \item \textbf{Decoupled and Robust Architecture}: We propose a novel three-stage decoupled framework—comprising shared visual feature extraction, layout analysis, and region-prompted decoding—integrated via a shared NaViT-based feature map. This design effectively mitigates the error propagation inherent in traditional sequential pipelines and the hallucinations common in end-to-end multimodal models, ensuring high structural and semantic fidelity.

    \item \textbf{High-Efficiency Dual-Track Parallelism}: We introduce a synergistic parallel decoding strategy that integrates \textit{Token Parallelism} and \textit{Query Parallelism}. Specifically, Token Parallelism achieves a lossless $5\text{--}11\times$ speedup through hybrid mask training, while Query Parallelism further accelerates the process by enabling concurrent region extraction. Collectively, these mechanisms yield a substantial enhancement in overall throughput compared to standard autoregressive decoding.
    
    \item \textbf{Comprehensive and Versatile Recognition}: We design Youtu-Parsing to recognize a diverse array of document elements, including text, formulas, tables, charts, seals, and hierarchical structures. This integrated approach allows us to process various real-world documents—such as academic publications, legal filings, and presentation slides.
    
    \item \textbf{State-of-the-Art Performance}: Extensive evaluations on authoritative benchmarks, including OmniDocBench and olmOCR-bench, demonstrate that Youtu-Parsing achieves state-of-the-art (SOTA) performance. It consistently outperforms both general-purpose vision-language models and specialized parsing systems in terms of recognition accuracy, structural integrity, and inference efficiency.
\end{itemize}

\section{Youtu-Parsing}

\subsection{Model Architecture}


\begin{figure}[htbp]
  \centering
  \includegraphics[width=0.95\textwidth]{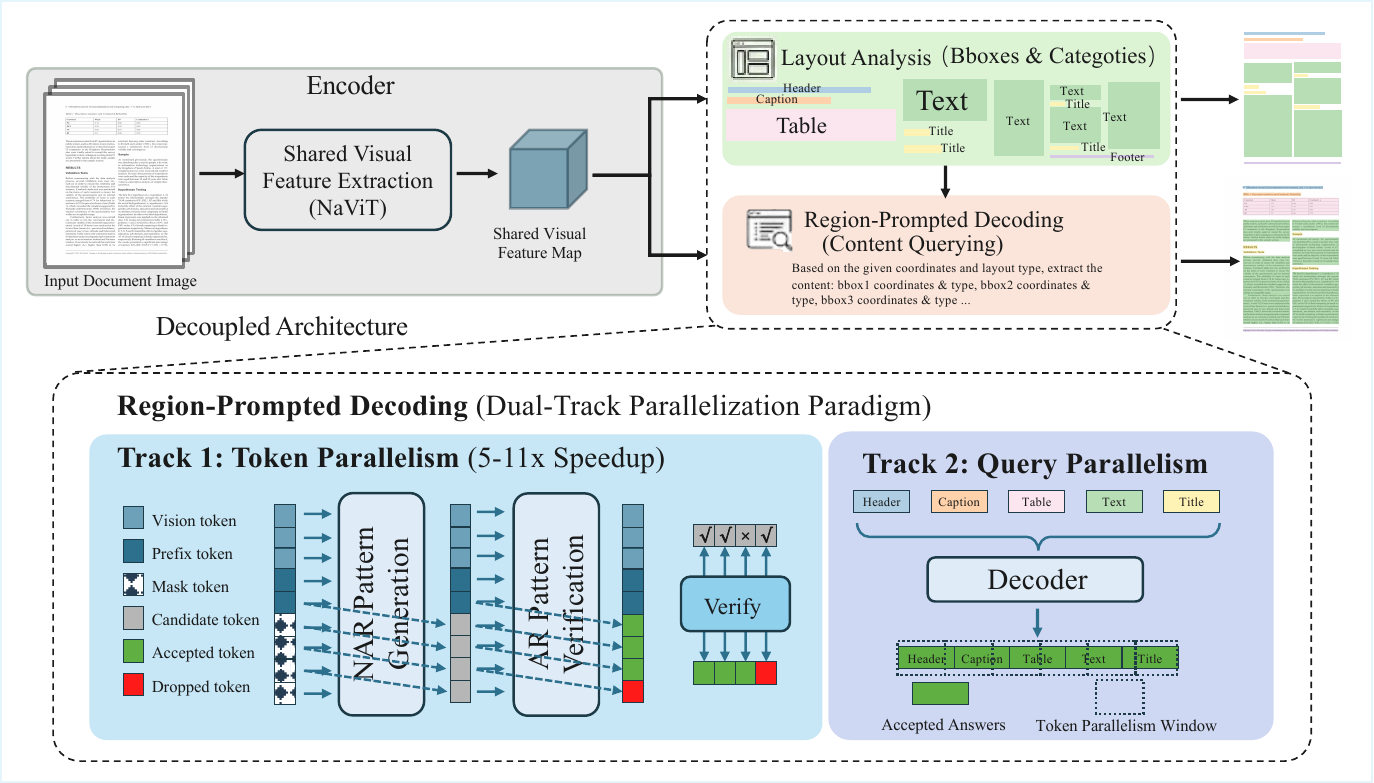}
  \caption{Overview of the proposed framework. The architecture utilizes a NaViT backbone for shared visual feature extraction, supporting both layout analysis and region-prompted decoding. We introduce two parallel mechanisms: \textbf{Track 1 (Token Parallelism)} for accelerated N-token prediction and verification ($5\times$ to $11\times$ speedup), and \textbf{Track 2 (Query Parallelism)} for concurrent multi-region extraction via category-specific prompts. }
  \label{fig:framework}
\end{figure}

Figure \ref{fig:framework} illustrates the overall architecture of Youtu-Parsing. The model structure is developed with reference to Youtu-VL \citep{youtu-vl}, integrating a large language model (LLM) with specialized visual modules to facilitate structured document understanding. Specifically, Youtu-Parsing employs Youtu-LLM-2B \citep{youtu-llm} as its core engine, leveraging its robust capacity for linguistic comprehension and generation. This is complemented by a pre-trained 0.4B-parameter vision encoder, equipped with a dynamic resolution preprocessor to extract high-level visual features from document images. To bridge the two modalities, a two-layer multi-layer perceptron (MLP) projector serves as the alignment module, mapping visual representations into the LLM's input space to enhance task-specific performance in document parsing.

\subsection{Parsing Strategy}

While end-to-end parsing methods are conceptually straightforward, they often entail substantial computational overhead, particularly when processing lengthy or complex documents. In contrast, element-level "crop-and-recognize" approaches are highly sensitive to the precision of layout coordinates; even marginal localization inaccuracies can precipitate recognition failures, thereby compromising downstream analysis. To address these limitations, Youtu-Parsing introduces an instruction-guided framework that prioritizes both structural integrity and computational efficiency. The pipeline consists of the following three stages:

\begin{itemize}
    \item \textbf{Shared Visual Feature Extraction:} The input document image is first processed by a NaViT-style Vision Transformer (ViT) encoder. This module encodes visual information into a shared, high-resolution feature map optimized for multi-scale representations, providing a consistent foundation for subsequent tasks.
    
    \item \textbf{Layout Analysis:} Guided by task-specific instruction prompts, the language model leverages global visual features to perform spatial localization and classification. This stage identifies the coordinates and semantic categories of various structural components, such as text blocks, tables, charts, and mathematical formulas.
    
    \item \textbf{Region-prompted Decoding:} For each localized element, a \textit{region-based prompt}—comprising spatial coordinates and category metadata—is utilized to retrieve targeted representations from the shared feature map. This mechanism enables the LLM to perform precise, fine-grained recognition of plain text, complex tabular structures, and LaTeX formulas. By decoupling localization from recognition, the framework ensures a highly efficient and reusable feature pipeline, facilitating the concurrent processing of heterogeneous document elements.
\end{itemize}
\subsection{Parallel Decoding Strategy}
\subsubsection{Token Parallelism}
\begin{figure}[htbp]
  \centering
  \includegraphics[width=0.85\textwidth]{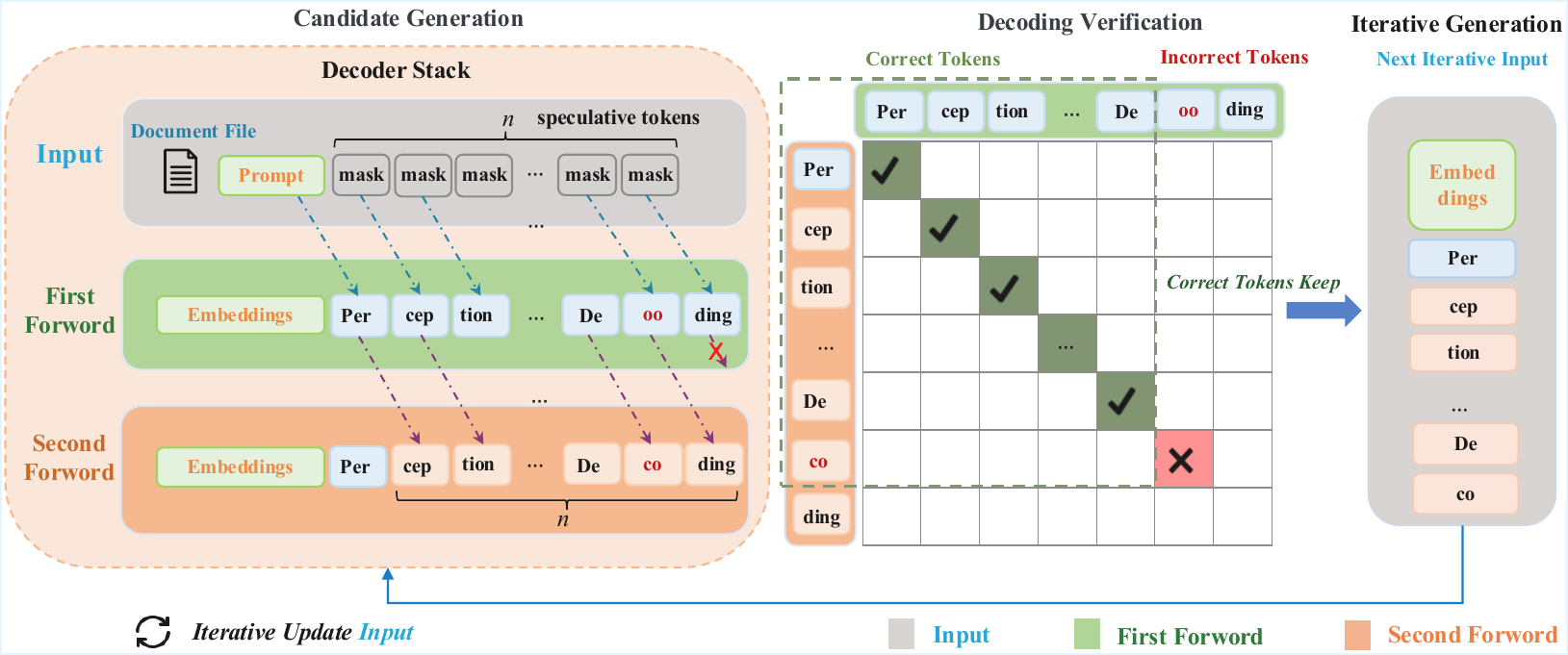}
  \caption{The framework of the parallel decoding strategy.}
  \label{fig:parallel_decoding}
\end{figure}

As illustrated in Figure~\ref{fig:parallel_decoding}, Youtu-Parsing incorporates an innovative parallel decoding mechanism that enhances standard autoregressive generation with speculative look-ahead capabilities via specialized mask tokens. Unlike conventional autoregressive models that generate tokens strictly sequentially, our approach concurrently predicts a batch of candidate tokens at each decoding step. These candidates are subsequently verified and filtered to ensure that the output remains mathematically identical to standard decoding. This strategy yields a practical throughput improvement of $5\text{--}11\times$ relative to standard methods, enabling the efficient processing of large-scale documents in real-time. The inference process is structured into three distinct stages:

\noindent\textbf{Candidate Generation.}
Given the current context sequence (comprising visual embeddings, system instructions, and the generated tokens thus far), we append $n$ special <mask> tokens (default $n=64$) to form an augmented input:
\begin{equation}
\mathbf{x}_{\text{input}} = [\text{context}] \oplus [\texttt{<mask\textgreater}]^n
\end{equation}
The model performs a single forward pass over this sequence to simultaneously predict: (a) the standard next token at the current autoregressive position, and (b) a set of $n$ candidate tokens at the subsequent mask positions.

\noindent\textbf{Decoding Verification.}
To maintain the lossless nature of the generation, we verify the speculative candidates through a second forward pass. We construct a verification sequence by concatenating the prompt with the candidates generated in the first stage. The model then performs an element-wise comparison between the mask-predicted candidates and the tokens that would have been generated autoregressively at those positions. We define the match at position $i$ as:
\begin{equation}
\text{match}_i = \mathbb{1}[\mathbf{x}_{\text{candidate}}[i] = \mathbf{x}_{\text{verified}}[i]]
\end{equation}
The verification logic identifies the first mismatch at position $k = \min\{i : \mathbf{x}_{\text{candidate}}[i] \neq \mathbf{x}_{\text{verified}}[i]\}$. Only the tokens preceding $k$ are accepted, ensuring the output is equivalent to sequential decoding.


\noindent\textbf{Iterative Generation.}
The accepted tokens are appended to the sequence, and the process iterates. In document parsing, the high predictability of structural patterns—such as LaTeX syntax and table formatting—consistently enables an average acceptance of $10\text{--}20$ tokens per iteration across various scenarios. Despite the overhead of two inference passes per step, this high average acceptance rate translates into a substantial $5\text{--}11\times$ speedup in token throughput compared to standard autoregressive decoding.

\noindent\textbf{Hybrid Masked Training.}
To empower the model with the look-ahead capabilities required for parallel decoding, we employ a Hybrid Masked Training (HMT) strategy during the fine-tuning phase. We adopt a dual-objective approach: 80\% of the training samples are augmented by inserting mask tokens at random positions with stochastic lengths. This masking scheme incentivizes the model to capture multi-token dependencies and internalize the structural regularities of document elements. Conversely, the remaining 20\% of the samples are processed using standard autoregressive training to preserve the model's baseline generation integrity. This hybrid paradigm ensures that Youtu-Parsing achieves high speculative accuracy without compromising its fundamental reasoning capabilities.

\subsubsection{Query Parallelism}

While Token Parallelism significantly accelerates the generation of individual content sequences, we observe that the default capacity of mask tokens ($n=64$) is often underutilized when processing short-form elements. Components such as headings, captions, labels, and single-line annotations are ubiquitous in real-world documents, yet their content length is typically far below the 64-token threshold. In such instances, the speculative decoding mechanism cannot fully exploit its parallelization potential, as a substantial portion of mask tokens remains idle, leading to computational redundancy and increased inference latency.

To mitigate this efficiency bottleneck, Youtu-Parsing introduces \textit{Query Parallelism}, a complementary strategy that enables the simultaneous extraction of multiple layout regions. By batching multiple region-prompted queries within a single forward pass, the model avoids the overhead of processing each layout element independently.

\noindent\textbf{Batched Query Construction.}
Given a set of layout elements $\{L_1, L_2, \ldots, L_m\}$ identified during the layout analysis stage—where each element $L_i$ is defined by its bounding box coordinates $(x_1^i, y_1^i, x_2^i, y_2^i)$ and semantic type $t_i$—we construct the batched input as follows:
\begin{equation}
\mathbf{Q}_{\text{batch}} = [\text{Instruction}] \oplus \bigoplus_{i=1}^{m} \left( \texttt{<x\_}x_1^i\texttt{>}\texttt{<y\_}y_1^i\texttt{>}\texttt{<x\_}x_2^i\texttt{>}\texttt{<y\_}y_2^i\texttt{>}\texttt{<LAYOUT\_}t_i\texttt{>} \mid \right)
\end{equation}
where the instruction prompt defines the extraction task (e.g., identifying content within specified regions, formatting formulas in LaTeX, and tables in OTSL \citep{lysak2023optimized}), and the delimiter "$\mid$" separates individual region queries. The maximum batch size is empirically set to $m=5$ to balance throughput gains against sequence length constraints.

\noindent\textbf{Efficiency Analysis.}
Query Parallelism is particularly effective for documents characterized by a high density of short text blocks. Consider a document page containing $k$ such blocks, each with an average length of $l$ tokens ($l \ll 64$). Under the conventional element-by-element paradigm, the model requires $k$ independent decoding sequences, each with a capacity utilization of only $l/64$. In contrast, Query Parallelism allows these $k$ elements to be processed in only $\lceil k/m \rceil$ batched passes.

For documents such as presentation slides, forms, and structured reports, Query Parallelism yields an approximate $2\times$ throughput improvement. This acceleration is synergistic with Token Parallelism: while the latter optimizes intra-sequence generation, Query Parallelism reduces inter-sequence overhead by amortizing the fixed costs of model invocation across multiple layout elements.

\noindent\textbf{Sequence Decomposition and Mapping.}
The model produces a unified output sequence containing the concatenated results of all queried regions, with individual segments demarcated by a specialized \texttt{<sep>} delimiter. To reconstruct the document structure, a post-processing module performs \textit{sequence decomposition} by partitioning the output based on these delimiters. Each extracted segment is then systematically mapped back to its corresponding layout element $\{L_i\}$ using the original query order. This deterministic association ensures that even when multiple regions are processed in parallel, the semantic and spatial integrity of the document is preserved for downstream reconstruction.

\subsection{Hierarchical Structure Analysis}

Complex documents—such as academic papers, technical reports, and presentation slides—exhibit intricate hierarchical structures encompassing sections, paragraphs, lists. Accurately parsing these relationships is pivotal for downstream applications, including structured information extraction, document summarization, and knowledge graph construction. 

Youtu-Parsing introduces a hierarchical structure analysis strategy based on specialized relational markers. This approach explicitly models the subordination, grouping, and continuation relationships among document elements by embedding three semantically distinct markers into the output sequence, enabling a high-fidelity representation of complex document topologies.

\begin{figure}[htbp]
  \centering
  \includegraphics[width=0.95\textwidth]{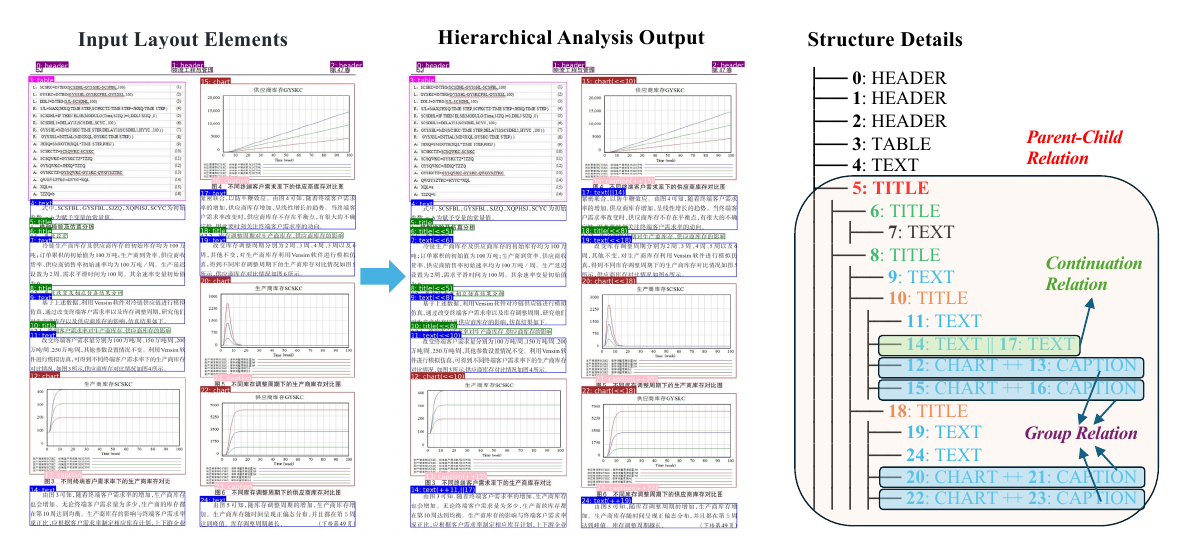}
  \caption{The framework of hierarchical structure analysis.}
  \label{fig:Hierarchical}
\end{figure}

\noindent\textbf{Relational Markup System.}
We define three specialized markers to characterize the logical associations between document elements:

\begin{itemize}
    \item \textbf{Parent-Child Relation (\text{<<}):} This marker denotes logical subordination. If element $B$ is a child node of element $A$ (e.g., a paragraph under a section heading, or a list item under a title), it is annotated as $A \text{ << } B$. This allows the model to explicitly reconstruct the document's tree-like taxonomy.
    
    \item \textbf{Grouping Relation (\text{++}):} This marker identifies logically associated sibling elements at the same hierarchical level. For a set of elements $\{E_1, E_2, \ldots, E_n\}$ sharing a common parent $T_1$, the structure is represented as $T_1 \text{ << } E_1$, followed by $E_2 \text{ ++ } E_1, E_3 \text{ ++ } E_2, \ldots, E_n \text{ ++ } E_{n-1}$. This chaining mechanism effectively captures lateral semantic associations.
    
    \item \textbf{Continuation Relation (\text{||}):} This marker addresses content fragmented by physical layout constraints, such as multi-column formatting or page breaks. When a single semantic unit $C$ is split into segments $C_1$ and $C_2$, the notation $C_1 \text{ || } C_2$ indicates that these segments should be merged during post-processing to restore textual integrity.
\end{itemize}

\noindent\textbf{Structural Inference and Representation.}
The hierarchical representation process is illustrated in Figure~\ref{fig:Hierarchical}. Given the elements identified during layout analysis, the model predicts the relationship type between element pairs by fusing visual features with hierarchical instruction prompts. The resulting output is a linearized sequence where relational markers encode the document's structural topology. Elements without explicit logical dependencies are excluded from this relational sequence to maintain conciseness.

\subsection{Training Recipe}

The training procedure follows a multi-stage curriculum, as summarized in Table \ref{tab:training-setting}. The overarching objective is to progressively evolve the model's capabilities—transitioning from broad-scale foundational pre-training to task-specific supervised fine-tuning (SFT), and ultimately to reinforcement learning refinement.

\begin{table}[htbp]
\caption{Training configurations across different stages.}
\label{tab:training-setting}
\centering
\begin{tabular}{lccc}
\toprule
\textbf{Parameters}  & \textbf{Stage 1: Pre-training} & \textbf{Stage 2: SFT} & \textbf{Stage 3: RL} \\
\midrule
Training Samples  & 30M & 3M & 20k \\
Max Resolution & $12288 \times 32 \times 32$ & $12288 \times 32 \times 32$ & $12288 \times 32 \times 32$  \\
Sequence Length & 20,480 & 20,480 & 20,480 \\
Trainable Components & All & All & All \\
Batch Size & 400 & 400 & 128 \\
Learning Rate   & $5 \times 10^{-5}$ & $5 \times 10^{-6}$ & $1 \times 10^{-6}$ \\
\bottomrule
\end{tabular}
\end{table}

\subsubsection{Stage 1: Pre-training}
The primary objective of this phase is to establish a robust foundation for document perception through large-scale supervised learning on 30M high-quality, OCR-centric samples. We train the model for 2 epochs with a learning rate of $5 \times 10^{-5}$ and a batch size of 400. To accommodate high-density information, we employ a sequence length of 20,480 and an input resolution of $12288 \times 32 \times 32$. Data augmentation techniques, including geometric transformations and noise injection, are applied to enhance perceptual robustness against low-quality or distorted document scans.

\subsubsection{Stage 2: Supervised Fine-tuning}
In the second stage, the model undergoes supervised fine-tuning (SFT) on a curated corpus of 3M expert-annotated documents. Maintaining the same resolution and sequence length as Stage 1, we reduce the learning rate to $5 \times 10^{-6}$ to facilitate stable adaptation. This stage focuses on five functional dimensions to achieve holistic document understanding:
\begin{itemize}
    \item \textbf{Layout Detection:} Precise spatial coordinate prediction for document elements to support reading order reconstruction.
    \item \textbf{Text Recognition:} Accurate textual extraction across heterogeneous fonts and complex background layouts.
    \item \textbf{Formula Recognition:} Mapping mathematical expressions and scientific notations to standardized LaTeX format.
    \item \textbf{Table Recognition:} Structural parsing using the Optimized Table-Structure Language (OTSL).
    \item \textbf{Chart Recognition:} Translating visual charts (e.g., bar, line, and pie charts) into structured tables or Mermaid notation.
\end{itemize}

\subsubsection{Stage 3: Reinforcement Learning}

To further align model outputs with structural constraints and human preferences, we employ Group Relative Policy Optimization (GRPO) \citep{shao2024deepseekmath} on 20k high-complexity samples. For each query $q$, a group of outputs $\{o_1, \dots, o_G\}$ is sampled from the old policy $\pi_{\theta_{\text{old}}}$. The optimization objective is formulated as:

\begin{equation}
\mathcal{J}_{\text{GRPO}}(\theta) = \mathbb{E}_{q \sim \mathcal{D}, \{o_i\}_{i=1}^G \sim \pi_{\theta_{\text{old}}}} \left[ \frac{1}{G} \sum_{i=1}^G \left( \min \left( \rho_i \hat{A}_i, \text{clip}(\rho_i, 1-\epsilon, 1+\epsilon) \hat{A}_i \right) - \beta \mathbb{D}_{\text{KL}}(\pi_\theta || \pi_{\text{ref}}) \right) \right]
\end{equation}

where $\rho_i = \frac{\pi_\theta(o_i|q)}{\pi_{\theta_{\text{old}}}(o_i|q)}$ is the probability ratio, and $\beta$ controls the KL-divergence penalty. Crucially, the advantage $\hat{A}_i$ is computed by normalizing the rewards within the generated group to reduce variance: $\hat{A}_i = \frac{r_i - \text{mean}(\{r_j\}_{j=1}^G)}{\text{std}(\{r_j\}_{j=1}^G)}$.

We define task-specific reward functions $r$ to provide fine-grained feedback across different modalities:
\begin{itemize}
    \item \textbf{Layout Analysis:} The reward evaluates syntactic compliance and geometric precision, quantified via optimal bipartite matching IoU between predicted and ground-truth bounding boxes.
    \item \textbf{Table Recognition:} We balance textual accuracy with topological fidelity by combining normalized Levenshtein similarity and Tree Edit Distance-based Similarity (TEDS), penalizing both content errors and structural hallucinations.
    \item \textbf{Formula Recognition:} To ensure syntactic and semantic integrity, the reward aggregates four dimensions: character edit distance, structural skeleton similarity (abstracting variables), symbol Jaccard overlap, and delimiter consistency.
\end{itemize}

\section{Data Processing}

\subsection{Open-source Data}

\begin{figure}[htbp]
  \centering
  \includegraphics[width=0.85\textwidth]{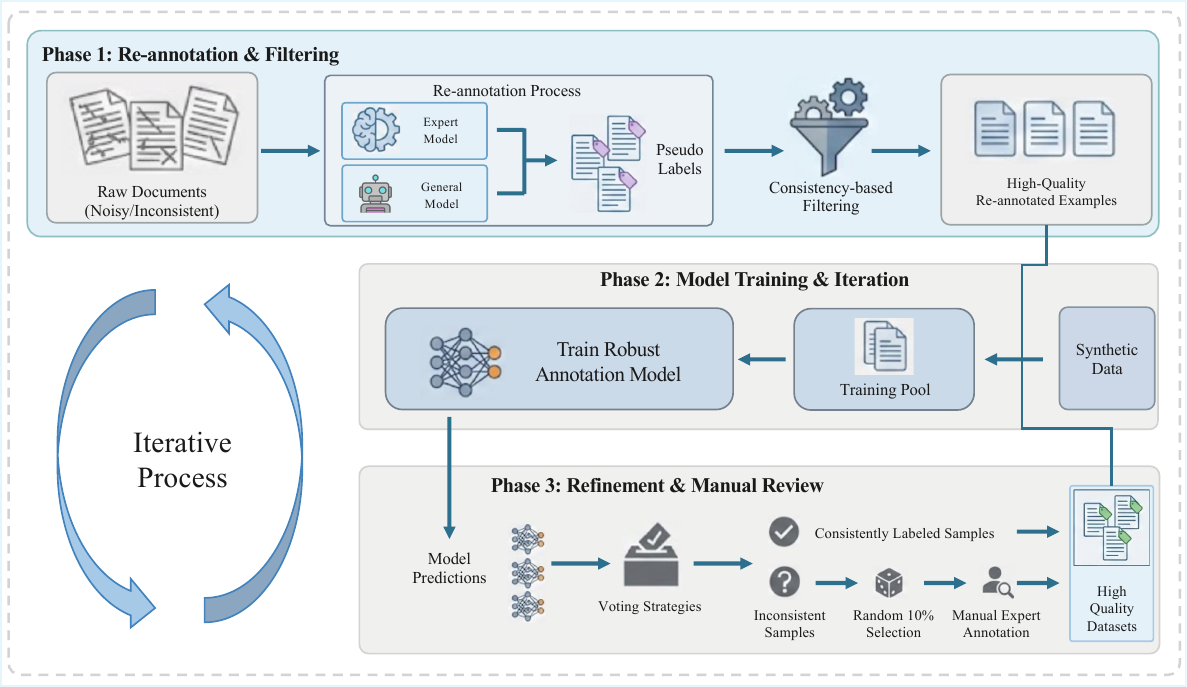}
  \caption{The iterative data refinement pipeline. Raw documents undergo re-annotation via specialized expert models and large language models (LLMs), followed by consistency-based filtering to ensure label fidelity. High-quality samples are then integrated with synthetic data to retrain the annotation model, facilitating further iterative refinement. A multi-model voting strategy and targeted manual verification are employed to resolve challenging cases, progressively building a large-scale, high-fidelity dataset.}
  \label{fig:data_processing1}
\end{figure}

While open-source datasets provide extensive diversity across document layouts and domains, their original annotations often exhibit significant noise and structural inconsistencies. To mitigate these limitations, we implement an iterative refinement pipeline that elevates annotation quality through a "model-in-the-loop" approach. The process begins by re-annotating raw open-source documents using a combination of specialized expert models and large language models (LLMs). Subsequently, we apply a consistency-based filtering mechanism to systematically prune unreliable pseudo-labels, ensuring that only high-fidelity annotations are retained for subsequent stages.

To further enhance the robustness of our training pool, these refined examples are augmented with high-quality synthetic data. This consolidated dataset is then used to retrain a more powerful annotation model, which is iteratively redeployed to refine the open-source corpus. To ensure label consensus, we introduce a multi-model voting strategy that aggregates predictions from various model architectures to identify stable, high-confidence samples. For ambiguous cases where model consensus cannot be reached, a 10\% subset is prioritized for expert manual verification. This iterative cycle—encompassing model-driven annotation, rigorous filtering, and targeted manual review—not only rectifies errors in open-source labels but also facilitates the continuous expansion of a diverse, high-quality training corpus, as illustrated in Figure \ref{fig:data_processing1}.

\subsection{Synthetic Data}
\begin{figure}[htbp]
  \centering
  \includegraphics[width=0.85\textwidth]{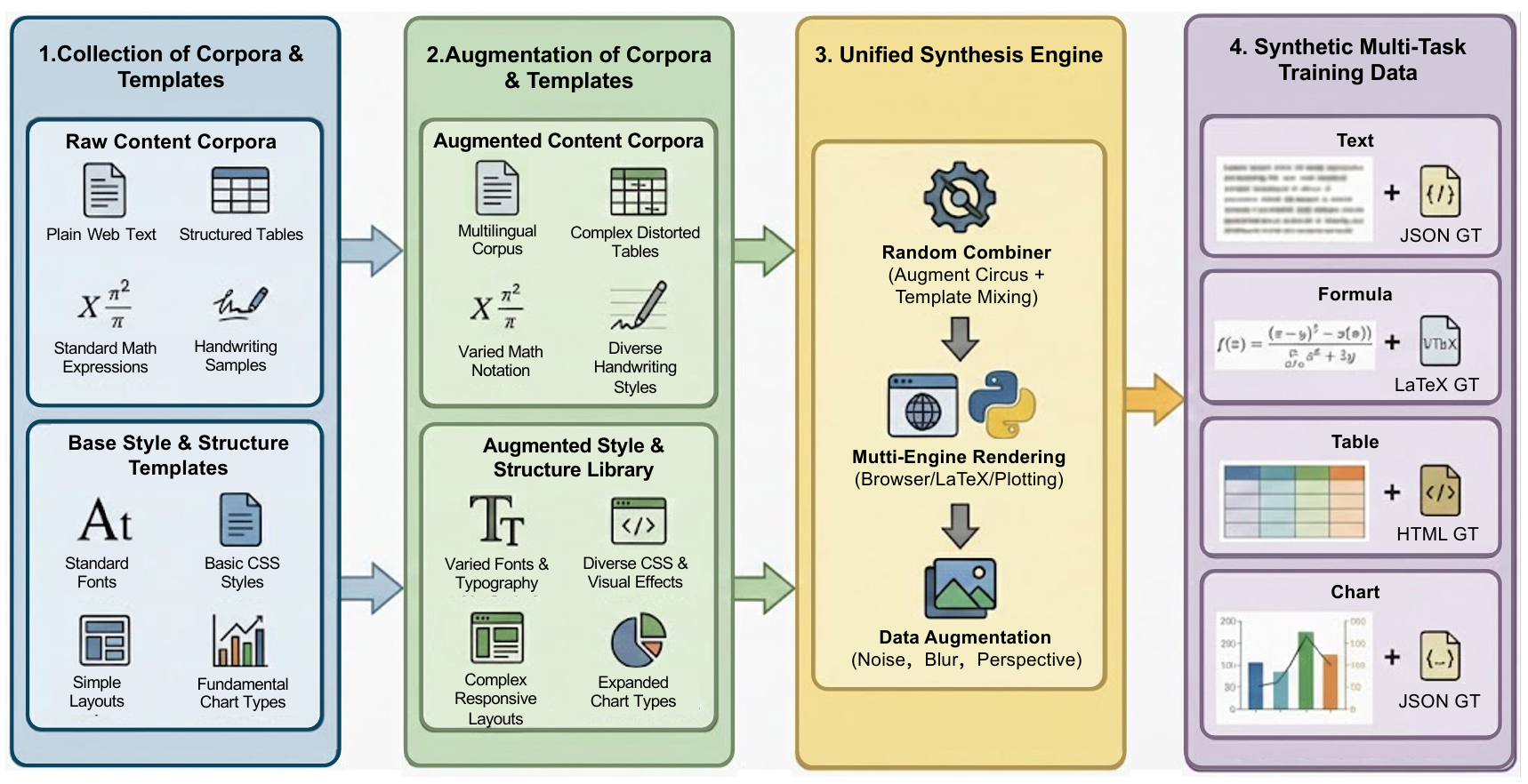}
  \caption{Overview of the unified synthetic data pipeline for document parsing, including four key stages: (1) Collection of Corpora \& Templates, (2) Augmentation of Corpora \& Templates, (3) Unified Synthesis Engine, and (4) Generation of Synthetic Multi-Task Data. Each stage contributes to creating diverse, annotated benchmark data for OCR, formula, table, and chart parsing tasks.}
  \label{fig:data_processing2}
\end{figure}

Synthetic data is a critical resource in advancing document intelligence, especially when large-scale annotated datasets are scarce, incomplete, or challenging to obtain. Through controlled data generation, synthetic datasets effectively address issues such as data scarcity, imbalanced distributions, and the need for enhanced model robustness and generalization across diverse document types. To this end, we propose a unified synthetic data pipeline for document parsing (as illustrated in Figure \ref{fig:data_processing2}).

\textbf{Collection of Corpora \& Templates.} The foundation of synthetic document generation lies in the systematic assembly of diverse content corpora and structural templates. This dual-source approach ensures that the synthetic datasets possess the necessary complexity and heterogeneity for robust model training.
\begin{itemize}
\item\textbf{Raw Content Corpora:} These encompass a wide spectrum of document elements, including multilingual text, rare characters, and handwritten content, alongside mathematical formulas, tables, and charts. This diverse collection ensures broad linguistic coverage and the ability to recognize low-frequency glyphs and varied writing styles.
\item\textbf{Base Style \& Structure Templates:} These consist of various document categories such as magazines, research articles, educational resources, instruction manuals, and advertisements. These templates provide a structural framework that encompasses representative layouts and formatting conventions for both printed and handwritten data synthesis.
\end{itemize}

\textbf{Augmentation of Corpora \& Templates.} To counteract data imbalance and expand dataset diversity, targeted augmentation is applied at both the content and template level. This process is instrumental in enhancing model resilience and generalization for complex, practical parsing scenarios.
\begin{itemize}
\item \textbf{Augmented Content Corpora:} Content is diversified by varying languages, formula symbols (LaTeX), table cell values, and chart data distributions.
\item \textbf{Augmented Style \& Structure Library:} Style templates are further enhanced through variations in document layouts, formula formatting, table structures, and chart visual attributes, such as color schemes and axis styles.
\end{itemize}

\textbf{Unified Synthesis Engine.} The synthesis engine integrates enriched corpora and augmented templates to generate diverse, realistic synthetic datasets.
\begin{itemize}
\item \textbf{Random Combiner:} Augmented content and templates are stochastically paired to produce unique document instances, ensuring broad representativeness.
\item \textbf{Multi-Engine Rendering:} Specialized rendering modules generate high-fidelity document images for text, formulas, tables, and charts.
\item \textbf{Data Augmentation:} A suite of transformations, such as geometric distortions, noise injection, and visual modifications, further increase dataset diversity and simulate real-world document artifacts.
\end{itemize}

\textbf{Synthetic Multi-Task Data.} The resulting synthetic dataset supports multiple document parsing tasks, each paired with rich, benchmark-style ground truth annotations.
\begin{itemize}
\item \textbf{Text:} Includes document images with bounding box coordinates and transcription labels in structured formats.
\item \textbf{Formula:} Provides images alongside canonical LaTeX representations to facilitate precise formula recognition and conversion.
\item \textbf{Table:} Consists of table images and cell-level metadata, capturing both layout and content information necessary for advanced table understanding.
\item \textbf{Chart:} Contains chart images, annotated graphical elements, and linked structured data tables to enhance the training and evaluation of automated chart parsing systems.
\end{itemize}

\section{Evaluation}

This section presents a comprehensive quantitative assessment of Youtu-Parsing, demonstrating its efficacy across a wide range of document parsing tasks. To provide a rigorous benchmark, we compare Youtu-Parsing against two primary categories of baseline models: \textbf{general-purpose vision-language models (VLMs)}, which possess broad multimodal capabilities, and \textbf{advanced specialized models}, which are specifically optimized for document-centric understanding and parsing pipelines. The evaluation is structured into two parts: Section \ref{sec:Complete Document Parsing} reports the model's end-to-end performance on full-document parsing benchmarks, while Section \ref{sec:Element Document Parsing} provides a granular analysis of its proficiency in parsing specific document elements, such as text, formulas, and tables.

\subsection{Comprehensive Document Parsing Evaluation}
\label{sec:Complete Document Parsing}

We evaluate the end-to-end document parsing capabilities of Youtu-Parsing on two prominent open-source benchmarks: OmniDocBench v1.5 \citep{ouyang2025omnidocbench} and olmOCR-bench \citep{poznanski2025olmocr}. These datasets cover a wide spectrum of document types and complex layouts, providing a rigorous testbed for holistic parsing performance. Our results indicate that Youtu-Parsing achieves robust and consistent performance across both benchmarks, significantly outperforming existing solutions in handling diverse document scenarios.

\subsubsection{OmniDocBench v1.5}

\begin{table}[htbp]
\setcitestyle{numbers, square}
\begin{center}
\resizebox{\linewidth}{!}{
\begin{tabular}{cllcccccc}
\toprule
\textbf{Model Type}  & \textbf{Methods} & \textbf{Parameters} & 
\makecell[c]{\textbf{Overall $\uparrow$}} & 
\makecell[c]{\textbf{Text$^{Edit} \downarrow$}} & 
\makecell[c]{\textbf{Formula$^{CDM} \uparrow$}} & 
\makecell[c]{\textbf{Table$^{TEDS} \uparrow$}} & 
\makecell[c]{\textbf{Table$^{TEDS_{s}} \uparrow$}} & 
\makecell[c]{\textbf{Reading Order$^{Edit} \downarrow$}} \\
\midrule 
\multirow{3}{*}{\textbf{Pipeline Tools}} 
& Marker-1.8.2 \citep{Marker} & - & 71.30 & 0.206 & 76.66 & 57.88 & 71.17 & 0.250 \\
& Mineru2-pipeline \citep{wang2024mineru} & - & 75.51 & 0.209 & 76.55 & 70.90 & 79.11 & 0.225 \\
& PP-StructureV3 \citep{cui2025paddleocr} & - & 86.73 & 0.073 & 85.79 & 81.68 & 89.48 & 0.073 \\
\midrule 
\multirow{5}{*}{\textbf{General VLMs}} 
& GPT-4o \citep{hurst2024gpt} & - & 75.02 & 0.217 & 79.70 & 67.07 & 76.09 & 0.148 \\
& InternVL3-76B \citep{zhu2025internvl3} & 76B & 80.33 & 0.131 & 83.42 & 70.64 & 77.74 & 0.113 \\
& InternVL3.5-241B \citep{wang2025internvl3} & 241B & 82.67 & 0.142 & 87.23 & 75.00 & 81.28 & 0.125 \\
& Qwen2.5-VL-72B \citep{Qwen2.5-VL} & 72B & 87.02 & 0.094 & 88.27 & 82.15 & 86.22 & 0.102 \\
& Gemini-2.5 Pro \citep{comanici2025gemini} & - & 88.03 & 0.075 & 85.82 & 85.71 & 90.29 & 0.097 \\
\midrule 
\multirow{13}{*}{\textbf{Specialized VLMs}} 
& Dolphin \citep{feng2025dolphin} & 322M & 74.67 & 0.125 & 67.85 & 68.70 & 77.77 & 0.124 \\
& OCRFlux-3B \citep{Ocrflux} & 3B & 74.82 & 0.193 & 68.03 & 75.75 & 80.23 & 0.202 \\
& Mistral OCR \citep{Mistralocr} & - & 78.83 & 0.164 & 82.84 & 70.03 & 78.04 & 0.144 \\
& POINTS-Reader \citep{liu2025points} & 3B & 80.98 & 0.134 & 79.20 & 77.13 & 81.66 & 0.145 \\
& olmOCR-7B \citep{poznanski2025olmocr} & 7B & 81.79 & 0.096 & 86.04 & 68.92 & 74.77 & 0.121 \\
& MinerU2-VLM \citep{wang2024mineru} & 0.9B & 85.56 & 0.078 & 80.95 & 83.54 & 87.66 & 0.086 \\
& Nanonets-OCR-s \citep{Nanonets-OCR-S} & 3B & 85.59 & 0.093 & 85.90 & 80.14 & 85.57 & 0.108 \\
& MonkeyOCR-pro-1.2B \citep{li2025monkeyocr} & 1.9B & 86.96 & 0.084 & 85.02 & 84.24 & 89.02 & 0.130 \\
& MonkeyOCR-3B \citep{li2025monkeyocr} & 3.7B & 87.13 & 0.075 & 87.45 & 81.39 & 85.92 & 0.129 \\
& dots.ocr \citep{li2025dots} & 3B & 88.41 & 0.048 & 83.22 & 86.78 & 90.62 & 0.053 \\
& MonkeyOCR-pro-3B \citep{li2025monkeyocr} & 3.7B & 88.85 & 0.075 & 87.25 & 86.78 & 90.63 & 0.128 \\
& MinerU2.5 \citep{niu2025mineru2} & 1.2B & 90.67 & 0.047 & 88.46 & 88.22 & 92.38 & 0.044 \\
& PaddleOCR-VL \citep{cui2025paddleocrvl} & 0.9B & 92.56 & 0.035 & 91.43 & 89.76 & 93.52 & 0.043 \\
& \cellcolor{red!20}\textbf{Youtu-Parsing} & \cellcolor{red!20}2.5B & \cellcolor{red!20}93.22 & \cellcolor{red!20}0.045 & \cellcolor{red!20}93.19 & \cellcolor{red!20}91.15 & \cellcolor{red!20}95.43 & \cellcolor{red!20}0.026 \\
\bottomrule 
\end{tabular}
}
\caption{Comprehensive evaluation of document parsing on OmniDocBench v1.5. $\uparrow$ denotes higher is better, $\downarrow$ denotes lower is better.}
\label{tab:OmniDocBench v1.5}
\end{center}
\end{table}

OmniDocBench v1.5 serves as a rigorous and extensive benchmark for the holistic evaluation of document parsing systems. It comprises a diverse collection of bilingual (Chinese and English) samples, with a specific focus on the structural complexity of mathematical formulas and layout elements. The evaluation framework utilizes the Character Detection Matching (CDM) \citep{Wang_2025_CVPR} for formula recognition, while the overall performance is quantified through a weighted aggregation of text, formula, and table-level metrics, offering a multi-dimensional assessment of parsing fidelity.

As summarized in Table \ref{tab:OmniDocBench v1.5}, \textbf{Youtu-Parsing} establishes a new state-of-the-art (SOTA) with an overall score of 93.22, consistently outperforming pipeline-based tools, general-purpose VLMs, and specialized parsing models. Specifically, the model demonstrates superior precision in both textual and structural reconstruction, as evidenced by its leading performance in Text-Edit distance (0.045), reading-order error (0.026), Formula-CDM (93.19), and Table-TEDS/TEDS-S (91.15/95.43). These results highlight its exceptional robustness in parsing complex mathematical expressions and preserving intricate tabular topologies, collectively validating the effectiveness and versatility of Youtu-Parsing for end-to-end document understanding.

\subsubsection{olmOCR-bench}

The olmOCR-bench is a specialized benchmark designed to evaluate document parsing models through unit test pass rates across a broad spectrum of real-world scenarios. The evaluation corpus consists of 1,403 PDF documents, spanning diverse categories such as arXiv preprints, historical scanned archives, complex tables, and multi-column layouts. Notably, the benchmark includes rigorous test cases targeting particularly challenging elements, such as microscopic text and intricate header/footer structures. Performance is quantified via pass rate metrics, with error margins incorporated to ensure statistical significance and reliability. By offering targeted and high-granularity evaluation cases, olmOCR-bench enables a robust and nuanced assessment of a model's parsing capabilities, particularly in handling edge-case scenarios.

\begin{table*}[htbp]
\setcitestyle{numbers, square}
\centering
\resizebox{\linewidth}{!}{
\begin{tabular}{lccccccccc}
\toprule
\textbf{Methods} & \multicolumn{9}{c}{\textbf{Unit Test Pass Rate $\uparrow$}} \\
\cmidrule(lr){2-10}
& \textbf{Overall} & \textbf{ArXiv} & \textbf{Old Scans Math} & \textbf{Tables} & \textbf{Old Scans} & \textbf{Headers and Footers} & \textbf{Multi column} & \textbf{Long Tiny Text} & \textbf{Base} \\
\midrule
GOT \citep{wei2024general} & $48.3 \pm 1.1$ & 52.7 & 52.0 & 0.2 & 22.1 & 93.6 & 42.0 & 29.9 & 94.0 \\
Gemini Flash 2 (No Anchor) \citep{team2025google} & $57.8 \pm 1.1$ & 32.1 & 56.3 & 61.4 & 27.8 & 48.0 & 58.7 & 84.4 & 94.0 \\
MinerU-pipeline \citep{wang2024mineru} & $61.5 \pm 1.1$ & 75.4 & 47.4 & 60.9 & 17.3 & 96.6 & 59.0 & 39.1 & 96.6 \\
Gemini Flash 2 (Anchored)  \citep{team2025google} & $63.8 \pm 1.2$ & 54.5 & 56.1 & 72.1 & 34.2 & 64.7 & 61.5 & 71.5 & 95.6 \\
Nanonets-OCR-s \citep{Nanonets-OCR-S} & $64.5 \pm 1.1$ & 67.0 & 68.6 & 77.7 & 39.5 & 40.7 & 69.9 & 53.4 & 99.3 \\
Qwen2.5-VL-7B (No Anchor) \citep{Qwen2.5-VL} & $65.5 \pm 1.2$ & 63.1 & 65.7 & 67.3 & 38.6 & 73.6 & 68.3 & 49.1 & 98.3 \\
GPT-4o (No Anchor) \citep{hurst2024gpt} & $68.9 \pm 1.1$ & 51.5 & 75.5 & 69.1 & 40.9 & 94.2 & 68.9 & 54.1 & 96.7 \\
GPT-4o (Anchored) \citep{hurst2024gpt} & $69.9 \pm 1.1$ & 53.5 & 74.5 & 70.0 & 40.7 & 93.8 & 69.3 & 60.6 & 96.8 \\
Marker-1.8.2 \citep{Marker} & $70.1 \pm 1.1$ & 76.0 & 57.9 & 57.6 & 27.8 & 84.9 & 72.9 & 84.6 & 99.1 \\
olmOCR v0.1.75 (No Anchor) \citep{poznanski2025olmocr} & $74.7 \pm 1.1$ & 71.5 & 71.4 & 71.4 & 42.8 & 94.1 & 77.7 & 71.0 & 97.8 \\
olmOCR v0.1.75 (Anchored) \citep{poznanski2025olmocr} & $75.5 \pm 1.0$ & 74.9 & 71.2 & 71.0 & 42.1 & 94.5 & 78.3 & 73.3 & 98.3 \\
MonkeyOCR-pro-3B \citep{li2025monkeyocr} & $75.8 \pm 1.0$ & 83.8 & 68.8 & 74.6 & 36.1 & 91.2 & 76.6 & 80.1 & 95.3 \\
MinerU2.5 \citep{niu2025mineru2} & $77.5 \pm 1.0$ & 81.1 & 74.0 & 85.1 & 33.8 & 96.3 & 65.5 & 89.8 & 94.4 \\
dots.ocr \citep{li2025dots} & $79.1 \pm 1.0$ & 82.1 & 64.2 & 88.3 & 40.9 & 94.1 & 82.4 & 81.2 & 99.5 \\
PaddleOCR-VL \citep{cui2025paddleocrvl} & $80.0 \pm 1.0$ & 85.7 & 71.0 & 84.1 & 37.8 & 97.0 & 79.9 & 85.7 & 98.5 \\
\cellcolor{red!20}\textbf{Youtu-Parsing} & \cellcolor{red!20}$80.5 \pm 0.9$ & \cellcolor{red!20}83.7 & \cellcolor{red!20}71.4 & \cellcolor{red!20}89.0 & \cellcolor{red!20}36.7 & \cellcolor{red!20}94.3 & \cellcolor{red!20}83.1 & \cellcolor{red!20}87.8 & \cellcolor{red!20}97.8 \\
\bottomrule
\end{tabular}
}
\caption{Comprehensive evaluation of document parsing on olmOCR-Bench. $\uparrow$ denotes higher is better.}
\label{tab:olm_ocr_bench}
\end{table*}

As summarized in Table \ref{tab:olm_ocr_bench}, \textbf{Youtu-Parsing} establishes a new state-of-the-art (SOTA) with an overall pass rate of $80.5 \pm 0.9$ on the olmOCR-bench, demonstrating superior generalization across a wide spectrum of document formats and structural layouts. Specifically, the model delivers leading performance in challenging sub-categories, notably achieving 89.0 on Tables, 83.1 on Multi-column layouts, and 87.8 on Long Tiny Text. It also maintains high fidelity in specialized domains such as ArXiv (83.7) and Old Scans Math (71.4), while achieving a robust score of 97.8 on the Base subset. These results underscore the model's proficiency in navigating high-difficulty scenarios, ranging from mathematical notation in degraded historical scans to intricate tabular topologies. The consistent excellence across these heterogeneous tasks validates the robustness and versatility of Youtu-Parsing for large-scale, real-world document digitization and understanding.

\subsection{Fine-Grained Document Parsing Evaluation}
\label{sec:Element Document Parsing}

This section provides a granular assessment of Youtu-Parsing's capabilities in fine-grained document parsing. We systematically evaluate its performance across five core sub-tasks: text recognition, table analysis, formula parsing, chart interpretation, and seal recognition. To ensure a balanced evaluation of both breadth and depth, we utilize a combination of established public benchmarks and curated in-house datasets that encompass diverse domains, complex layouts, and varying imaging conditions. This rigorous experimental framework facilitates a thorough examination of the robustness, generalizability, and practical utility of Youtu-Parsing across a wide array of real-world document scenarios.

\subsubsection{Text Recognition}
We assess the text recognition capability of Youtu-Parsing using an in-house dataset designed to cover four representative scenarios: 

\begin{itemize}
\item \textbf{Handwritten Text:} consists of 10,721 images derived from sources such as essays, memos, notes, and various handwritten documents.
\item \textbf{Vertical Text:} comprises 3,865 images featuring content like advertisements, posters, and banners.
\item \textbf{Artistic Text:} includes 8,512 images encompassing text logos, greeting cards, and home decorations.
\item \textbf{High-resolution Images:} contains 9,127 images, covering materials such as newspapers and subway signage.
\end{itemize}

\begin{table}[h]
\setcitestyle{numbers, square} 
\footnotesize
\centering
\begin{tabular}{lcccc}
\toprule
Method & Handwritten & Vertical Text & Artistic Text & High-resolution \\
\midrule
dots.ocr \citep{li2025dots} & 98.27 & 89.22 & 97.15 & 92.32  \\
MinerU2.5 \citep{niu2025mineru2} & 86.29 & 71.28 & 92.10 & 95.48  \\
PaddleOCR-VL \citep{cui2025paddleocrvl} & 97.78 & 80.28 & 93.60 & 97.14  \\
\rowcolor{red!20} Youtu-Parsing & 98.94 & 89.91 & 97.36 & 97.38  \\
\bottomrule
\end{tabular}
\caption{Comparison of text recognition performance on the in-house OCR dataset.}
\label{tab:Inhouse_text_Edit}
\end{table}

We evaluate performance using the average edit distance similarity score (higher is better), as detailed in Table~\ref{tab:Inhouse_text_Edit}. Youtu-Parsing consistently achieves leading results across all scenarios. Specifically, its recognition accuracies for handwritten text, vertical text, artistic font, and high-resolution image subsets reach 98.94, 89.91, 97.36, and 97.38, respectively, outperforming all baseline methods. These results demonstrate the exceptional robustness and adaptability of Youtu-Parsing for complex text recognition tasks.

\subsubsection{Table Recognition}
We evaluate the table recognition capability of Youtu-Parsing on three public competition benchmarks: CC-OCR \citep{yang2025cc}, OCRBench v2 \citep{fu2024ocrbench}, and Inhouse Data as shown in Table \ref{tab:table_teds}. Below is an introduction to the three benchmarks:

\paragraph{CC-OCR} is a comprehensive multimodal OCR benchmark for the evaluation of large multimodal models (LMMs), collaboratively developed by Alibaba and several academic institutions. The dataset comprises 7,058 fully annotated images distributed across 39 subsets, with 41\% sourced from real-world scenarios. It covers four primary tracks: multi-scene and multilingual text reading, document parsing, and key information extraction. For our research, we retained data relevant to document recognition and specifically those images containing tables. Following these filtering steps, we obtained a subset of 300 images focused on table recognition.

\paragraph{OCRBench v2} is an enhanced, large-scale bilingual benchmark designed for evaluating the localization and reasoning capabilities of large multimodal models (LMMs) in optical character recognition (OCR) tasks. The benchmark comprises 23 distinct tasks spanning 8 core competencies, across 31 different scenarios, and contains 10,000 human-verified question-answer pairs as well as a private test set of 1,500 images. For our study, we selected data pertaining specifically to table recognition. After applying these filtering criteria, we curated a subset of 590 images dedicated to table recognition.

\paragraph{In-house Benchmark} is a custom dataset developed by ourselves to facilitate a more comprehensive evaluation of the table recognition capabilities of Youtu-Parsing. It comprises 800 images featuring a diverse range of document tables collected from various sources. To ensure broad coverage and challenge the model across different levels of complexity, the dataset includes multiple table types, such as simple tables, nested tables, and complex tables containing merged cells and irregular layouts.

\begin{table}[h]
\setcitestyle{numbers, square} 
\centering
\small
\begin{tabular}{lcccccc}
\toprule
\multirow{2}{*}{Method} & \multicolumn{2}{c}{CC-OCR} & \multicolumn{2}{c}{ OCRBench v2} & \multicolumn{2}{c}{Inhouse Data} \\
\cmidrule(lr){2-3} \cmidrule(lr){4-5} \cmidrule(lr){6-7}
& TEDS & TEDS-S & TEDS & TEDS-S & TEDS & TEDS-S  \\
\midrule
Qwen3-VL-2B \citep{qwen3technicalreport} & 54.70 & 60.42 & 50.93 & 57.41 & 62.54 & 65.42 \\
Qwen3-VL-4B \citep{qwen3technicalreport} & 58.00 & 62.71 & 61.62 & 66.38 & 68.00 & 71.62 \\
Qwen3-VL-8B \citep{qwen3technicalreport} & 60.32 & 64.80 & 65.78 & 70.95 & 70.95 & 74.54 \\
PaddleOCR-VL \citep{cui2025paddleocrvl} & 80.46 & 85.32 & 66.51 & 71.78 & 87.62 & 90.99 \\
\rowcolor{red!20} \textbf{Youtu-Parsing} & 81.37 & 86.04 & 72.55 & 78.31 & 88.24 & 91.63 \\
\bottomrule
\end{tabular}
\caption{Comparison of Table Edit Distance Recognition Performance.}
\label{tab:table_teds}
\end{table}

For the evaluation of table recognition, we specifically select images containing tables from the three benchmarks and conduct assessments using an end-to-end document parsing approach. As shown in Table \ref{tab:table_teds}, Youtu-Parsing achieves the highest scores across all datasets in both TEDS and TEDS-S metrics. Specifically, on the CC-OCR dataset, it obtains a TEDS of 81.37 and a TEDS-S of 86.04. For OCRBench v2, it reaches a TEDS of 72.55 and a TEDS-S of 78.31. On the Inhouse Data, it records a TEDS of 88.24 and a TEDS-S of 91.63. These results demonstrate that Youtu-Parsing is highly effective and reliable for end-to-end table parsing in practical scenarios.

\subsubsection{Formula Recognition}
We evaluate our model’s performance on the OmniDocBench v1.5 Formula and proprietary Inhouse Formula datasets using the Character Detection Matching (CDM) metric \citep{Wang_2025_CVPR}. The OmniDocBench v1.5 Formula evaluation set is sampled from OmniDocBench v1.5, while the Inhouse Formula dataset comprises 6,863 samples spanning diverse scenarios such as academic papers, textbooks across disciplines, and exam papers for different grade levels. As shown in Table~\ref{tab:Formula_recog}, our model achieves state-of-the-art CDM scores on both datasets, demonstrating the robust formula recognition capability of Youtu-Parsing in complex real-world scenarios.

\begin{table}[h]
\setcitestyle{numbers, square} 
\small
\centering
\begin{tabular}{lcc}
\toprule
Method & OmniDocBench v1.5 Formula & Inhouse Formula \\
\midrule
dots.ocr \citep{li2025dots} & 83.22 & 79.7  \\
MinerU2.5 \citep{niu2025mineru2} & 88.46 & 75.6  \\
PaddleOCR-VL \citep{cui2025paddleocrvl} & 91.43 & 75.0  \\
\rowcolor{red!20} \textbf{Youtu-Parsing} & 92.30 & 80.1  \\
\bottomrule
\end{tabular}
\caption{Comparison of Formula Recognition Performance.}
\label{tab:Formula_recog}
\end{table}

\subsubsection{Chart Recognition}

To evaluate the model's proficiency in chart understanding, we curated a comprehensive dataset of 1,000 real-world samples, categorized into two primary types: (1) \textbf{Data Charts} (500 samples), which include statistical visualizations such as bar, line, pie, radar, and area charts focused on numerical data extraction; and (2) \textbf{Logic Diagrams} (500 samples), which encompass relational structures such as flowcharts, mind maps, and architecture diagrams focused on structural parsing.

\begin{table}[htbp]
\setcitestyle{numbers, square} 
\small
\centering
\caption{Comparison of chart recognition performance. The \textbf{Data Charts} columns report accuracy in numerical data extraction (CSS and RMS-F1), while the \textbf{Logic Diagrams} column assesses structural parsing (Edge-F1).}
\label{tab:results_chart}
\begin{tabular}{lccc}
\toprule
\multirow{2}{*}{\textbf{Methods}} & \multicolumn{2}{c}{\textbf{Data Charts}} & \textbf{Logic Diagrams} \\ \cmidrule(lr){2-3} \cmidrule(l){4-4} 
 & \textbf{CSS $\uparrow$}  & \textbf{RMS-F1 $\uparrow$} & \textbf{Edge-F1 $\uparrow$} \\ \midrule
OneChart \citep{chen2024onechart} & 0.4783 & 0.2004 & - \\
Qwen3-VL-2B \citep{qwen3technicalreport} & 0.7467 & 0.3357 & 0.4780 \\
Qwen3-VL-4B \citep{qwen3technicalreport} & 0.8478 & 0.5204 & 0.4714 \\
Qwen3-VL-8B \citep{qwen3technicalreport} & 0.9076 & 0.5825 & 0.6484 \\
Qwen3-VL-32B \citep{qwen3technicalreport} & 0.9084 & 0.5925 & 0.7628 \\
PaddleOCR-VL \citep{cui2025paddleocrvl} & 0.8440 & 0.5316 & - \\
\rowcolor{red!20} \textbf{Youtu-Parsing} & 0.8995 & 0.6124 & 0.7699 \\ 
\bottomrule
\end{tabular}
\end{table}

To evaluate the conversion of \textbf{Data Charts} to Markdown format, we employ the standard RMS-F1 \citep{liu2023deplot} and our proposed \textbf{Content Similarity Score (CSS)}. While RMS-F1 is a common metric for chart-to-table tasks, it often suffers from high threshold sensitivity and a lack of structural awareness. CSS addresses these issues by providing a more robust measure of tabular data fidelity.

\paragraph{Content Similarity Score (CSS).} To assess semantic correctness while remaining invariant to structural permutations (e.g., row/column swaps), CSS parses the ground-truth table $\mathbf{A}$ and predicted table $\mathbf{P}$ into cell matrices. We first determine the optimal orientation by calculating the Levenshtein distance between the primary headers of $\mathbf{A}$ and $\mathbf{P}$ (including its transpose). Subsequently, a similarity matrix is constructed, and greedy bipartite matching is performed to establish the optimal alignment. Let $\hat{\mathbf{P}}$ denote the aligned prediction matrix. The cell-level similarity $s(a, p)$ is defined as:
\begin{equation}
s(a, p) = \begin{cases} 
1, & a = p, \\ 
0, & a = \varnothing \text{ or } p = \varnothing, \\ 
1 - \dfrac{d_{\text{edit}}(a, p)}{\max(|a|, |p|)}, & \text{otherwise}, 
\end{cases}
\end{equation}
where $d_{\text{edit}}(\cdot, \cdot)$ denotes the character-level edit distance. The final CSS is the average similarity across all data cells $\Omega$ (excluding headers):
\begin{equation}
\text{CSS} = \frac{1}{|\Omega|} \sum_{(i,j) \in \Omega} s\bigl(A_{ij}, \hat{P}_{ij}\bigr),
\end{equation}
where $\Omega = \{(i,j) \mid i \in [2, R],\, j \in [1, C]\}$ represents the data cell indices, with $R$ and $C$ being the maximum dimensions.

\paragraph{Edge-based F1 Score (Edge-F1).} To evaluate the structural integrity of \textbf{Logic Diagrams} converted to Mermaid syntax, we propose the \textbf{Edge-based F1 Score}. Given the ground-truth graph $G_a$ and predicted graph $G_p$, we extract their directed edge sets $E_a$ and $E_p$, where each edge is a tuple of normalized node labels. Precision ($P$), Recall ($R$), and the Edge-F1 score are defined as:
\begin{equation}
P = \frac{|E_a \cap E_p|}{\max(1, |E_p|)}, \quad 
R = \frac{|E_a \cap E_p|}{\max(1, |E_a|)}, \quad 
\text{Edge-}F1 = \frac{2 P R}{P + R}.
\end{equation}

As summarized in Table \ref{tab:results_chart}, \textbf{Youtu-Parsing} achieves the highest RMS-F1 (0.6124) and Edge-F1 (0.7699), demonstrating its superior capability in both numerical data reconstruction and structural reasoning. Notably, while its CSS (0.8995) is slightly lower than that of the much larger Qwen3-VL-32B (0.9084), Youtu-Parsing maintains highly competitive performance with significantly fewer parameters, highlighting its efficiency and robustness across diverse chart and diagram types.

\subsubsection{Seal Recognition}
To ensure the accuracy of our experimental results, we curated a comprehensive dataset comprising 500 real-world seals and 500 synthetic seals. This dataset covers various shapes—including oval, circular, square, diamond, and triangular—and incorporates multiple colors such as red, blue, and black, closely simulating real-world scenarios.

We evaluate performance using the average edit distance similarity score (higher is better). As presented in Table~\ref{tab:results_seal}, our model achieves outstanding accuracy, demonstrating that Youtu-Parsing delivers state-of-the-art seal recognition performance in complex real-world conditions.

\begin{table}[h]
  \setcitestyle{numbers, square} 
	\small
	\centering
	\begin{tabular}{lc}
		\toprule
		Method & Edit Similarity $\uparrow$ \\
		\midrule
		Qwen3-VL-2B \citep{qwen3technicalreport} & 60.83  \\
		Qwen3-VL-4B \citep{qwen3technicalreport} & 64.11  \\
		Qwen3-VL-8B \citep{qwen3technicalreport} & 70.90  \\
		Qwen3-VL-32B \citep{qwen3technicalreport} & 77.88  \\
		\rowcolor{red!20} \textbf{Youtu-Parsing} & 80.13  \\
		\bottomrule
	\end{tabular}
	\caption{Comparison of Seal Recognition Performance.}
	\label{tab:results_seal}
\end{table}

\subsection{Inference Performance}

Beyond parsing accuracy, computational efficiency is a pivotal factor for the practical deployment of document parsing systems. In this subsection, we conduct a comprehensive evaluation of Youtu-Parsing's inference performance, examining the efficacy of our proposed parallel decoding strategy under various parallelism configurations and attention mechanisms. Furthermore, we benchmark the end-to-end processing latency across diverse real-world scenarios.

\subsubsection{Token Parallelism Analysis}

To systematically investigate the impact of our parallel decoding strategy on both parsing quality and inference throughput, we evaluate Youtu-Parsing under different parallelism degrees, defined by the number of mask tokens $n$ appended during each decoding iteration. Additionally, we assess the synergy between the parallel decoding mechanism and two widely adopted attention implementations: \textbf{Flash Attention} and \textbf{Eager Attention} (standard scaled dot-product attention). Table~\ref{tab:parallel_decoding} summarizes the results on OmniDocBench v1.5, reporting the overall parsing score alongside the average number of accepted tokens per iteration—a key metric for parallelism efficiency. For all experiments, the Query Parallelism batch size was fixed at $m=5$.

\begin{table}[htbp]
\centering
\caption{Ablation study of parallel decoding with varying parallelism degrees ($n$) and attention mechanisms on OmniDocBench v1.5. "Avg. Accepted" denotes the average number of tokens accepted per decoding step; "Overall" represents the composite parsing score.}
\label{tab:parallel_decoding}
\small
\begin{tabular}{lcccc}
\toprule
\textbf{Attention Type} & \textbf{Token Parallelism ($n$)} & \textbf{Overall $\uparrow$} & \textbf{Avg. Accepted $\uparrow$} & \textbf{Speedup $\uparrow$} \\
\midrule
\multirow{3}{*}{Eager Attention} 
& 16 & 92.90 & 11 & 5.01$\times$ \\
& 32 & 92.90 & 16 & 6.92$\times$ \\
& 64 & 92.90 & 18 & 10.58$\times$ \\
\midrule
\multirow{3}{*}{Flash Attention} 
& 16 & 92.93 & 12 & 5.37$\times$ \\
& 32 & 93.17 & 18 & 7.54$\times$ \\
& 64 & 93.30 & 20 & 11.13$\times$ \\
\bottomrule
\end{tabular}
\end{table}

As shown in Table~\ref{tab:parallel_decoding}, increasing the parallelism degree $n$ consistently enhances the average number of accepted tokens per iteration, leading to substantial acceleration without compromising parsing quality. Additionally, even at the highest setting ($n=64$), Youtu-Parsing maintains performance on par with standard autoregressive decoding, validating the effectiveness of our two-pass verification mechanism in preserving output fidelity. Flash Attention achieves superior speedup ratios compared to Eager Attention while maintaining identical or slightly better accuracy, owing to its memory-efficient computation and optimized GPU utilization. These results demonstrate that our parallel decoding strategy effectively leverages modern attention optimizations for latency-sensitive applications.

To further evaluate the versatility of the Token Parallelism strategy, we measured inference latency and speedup ratios across four representative scenarios: \textbf{Table}, \textbf{Formula}, \textbf{Chart}, and \textbf{Text}. These scenarios exhibit distinct characteristics in terms of sequence length and structural complexity. The results are detailed in Table~\ref{tab:token-parallelism-scenarios}.

\begin{table}[htbp]
\caption{Token Parallelism performance across diverse document parsing scenarios. "Latency" indicates the average inference time per page (in seconds); "Speedup" is relative to the baseline autoregressive decoding ($n=1$).}
\label{tab:token-parallelism-scenarios}
\centering
\small
\begin{tabular}{lccc}
\toprule
\textbf{Scenario} & \textbf{Token Parallelism ($n$)} & \textbf{Latency (s) $\downarrow$} &  \textbf{Speedup $\uparrow$} \\
\midrule
\multirow{4}{*}{Table} 
  & 64  & 1.50   & 26.82$\times$ \\
  & 32  & 2.72  & 14.79$\times$ \\
  & 16  & 4.82  & 8.35$\times$ \\
  & 1   & 40.23  & 1.00$\times$ \\
\midrule
\multirow{4}{*}{Formula} 
  & 64  & 3.40   & 16.76$\times$ \\
  & 32  & 8.56   & 6.66$\times$ \\
  & 16  & 13.47   & 4.23$\times$ \\
  & 1   & 57.00  & 1.00$\times$ \\
\midrule
\multirow{4}{*}{Chart} 
  & 64  & 1.63   & 10.12$\times$ \\
  & 32  & 3.25   & 5.08$\times$ \\
  & 16  & 4.52   & 3.65$\times$ \\
  & 1   & 16.50   & 1.00$\times$ \\
\midrule
\multirow{4}{*}{Text} 
  & 64  & 18.37   & 13.61$\times$ \\
  & 32  & 37.80   & 6.62$\times$ \\
  & 16  & 65.50   & 3.81$\times$ \\
  & 1   & 250.05   & 1.00$\times$ \\
\bottomrule
\end{tabular}
\end{table}

The results yield several key observations. First, Token Parallelism provides significant speedup across all scenarios, with higher $n$ consistently reducing latency. Second, \textbf{structured output scenarios, particularly tables, exhibit the most pronounced acceleration}, achieving up to a 26.82$\times$ speedup at $n=64$. This is attributed to the high regularity and predictability of tabular content (e.g., repetitive delimiters and cell boundaries), which allows the model to successfully verify longer candidate sequences per iteration. In contrast, semantically dense content like free-form text shows relatively lower but still substantial gains. These findings underscore the efficacy of Token Parallelism in exploiting the structural redundancy inherent in document parsing, making it highly advantageous for large-scale digitization.

\subsubsection{Query Parallelism Analysis}

To evaluate the impact of the Query Parallelism strategy on both parsing performance and inference efficiency, we systematically evaluate the model across varying Query Parallelism degrees ($m \in \{1, 2, 3, 4, 5\}$) while maintaining a constant Token Parallelism degree ($n=64$). These experiments were conducted on our internal benchmark, with the results summarized in Table~\ref{tab:query-parallelism}.

The results demonstrate that increasing the Query Parallelism degree $m$ consistently reduces the average inference latency per page, confirming the efficacy of query batching in enhancing throughput. Specifically, as $m$ increases from 1 (sequential element processing) to 5, the average latency decreases by approximately 52\%, achieving a $2.09\times$ speedup. Notably, the parsing accuracy (Overall score) remains stable or slightly improves as $m$ increases, indicating that batching multiple layout queries does not compromise the model's recognition capabilities.

\begin{table}[htbp]
\caption{Impact of Query Parallelism on parsing performance and inference efficiency ($n=64$). "Query Parallelism ($m$)" denotes the number of layout elements processed per forward pass; "Latency" refers to the average processing time per page; "Speedup" is relative to sequential processing ($m=1$).}
\label{tab:query-parallelism}
\centering
\small
\begin{tabular}{cccc}
\toprule
\textbf{Query Parallelism ($m$)} & \textbf{Overall $\uparrow$} & \textbf{Latency (s/page) $\downarrow$} & \textbf{Speedup $\uparrow$} \\
\midrule
1 & 88.77 & 18.26 & $1.00\times$ \\
2 & 88.82 & 12.39 & $1.47\times$ \\
3 & 90.05 & 10.22 & $1.79\times$ \\
4 & 90.02 & 9.61 & $1.90\times$ \\
5 & \textbf{90.12} & \textbf{8.74} & \textbf{2.09}$\times$ \\
\bottomrule
\end{tabular}
\end{table}

In summary, the Query Parallelism strategy effectively mitigates inter-sequence processing overhead by batching multiple layout element queries into a single forward pass. This approach is particularly advantageous for documents characterized by a high density of short-text elements. When integrated, Token Parallelism and Query Parallelism function as complementary acceleration mechanisms: Token Parallelism optimizes \textbf{intra-sequence} generation efficiency, while Query Parallelism minimizes \textbf{inter-sequence} invocation overhead. Together, they achieve a substantial reduction in end-to-end inference latency, facilitating large-scale document processing.

\subsubsection{End-to-End Latency Comparison}

To contextualize the efficiency gains of Youtu-Parsing relative to existing document parsing solutions, we report the end-to-end inference latency across several representative methods. For each method, we measure the average processing time per document page on the OmniDocBench v1.5 test set under identical hardware configurations. Table~\ref{tab:latency_comparison} presents the comparative results.

\begin{table}[htbp]
\centering
\caption{End-to-end inference latency comparison on OmniDocBench v1.5. "Latency" denotes the average processing time per document page (in seconds); "Throughput" indicates the number of tokens generated per second.}
\label{tab:latency_comparison}
\setcitestyle{numbers, square} 
\small
\begin{tabular}{lccc}
\toprule
\textbf{Method} & \textbf{Parameters} & \textbf{Latency (s/page) $\downarrow$} & \textbf{Throughput (token/s) $\uparrow$} \\
\midrule
dots.ocr \citep{li2025dots} & 3B & 3.76 & 227 \\
MinerU2.5 \citep{niu2025mineru2} & 1.2B & 2.40 & 465 \\
PaddleOCR-VL \citep{cui2025paddleocrvl} & 0.9B & 1.48 & 1300 \\
\rowcolor{red!20} \textbf{Youtu-Parsing} & 2.5B & 1.75 & 445 \\
\bottomrule
\end{tabular}
\end{table}

The results presented in Table~\ref{tab:latency_comparison} demonstrate that Youtu-Parsing achieves highly competitive inference efficiency despite its substantially larger model capacity. While Youtu-Parsing comprises 2.5 billion parameters—significantly exceeding the scale of most compared methods—the integration of our parallel decoding strategy enables it to attain per-page latency comparable to, or even lower than, lighter-weight alternatives. This result underscores the efficacy of our parallel decoding mechanism in bridging the gap between model capacity and inference efficiency: a larger model does not necessarily entail proportionally higher latency. Combined with its state-of-the-art parsing accuracy demonstrated in preceding sections, Youtu-Parsing offers a compelling balance between performance and throughput, positioning it as a highly practical solution for large-scale document digitization and real-time document understanding applications.

\section{Conclusion}

This paper presents \textbf{Youtu-Parsing}, a 2.5-billion-parameter vision-language model designed to advance document parsing via a decoupled, prompt-guided architecture. To overcome the inherent latency of traditional autoregressive decoding, we introduce a dual-track high-parallelism strategy—integrating \textit{Token Parallelism} and \textit{Query Parallelism}—inspired by non-autoregressive paradigms. This framework achieves a substantial 10--20$\times$ throughput enhancement while ensuring bit-level equivalence to standard autoregressive outputs, thereby reconciling the long-standing trade-off between inference speed and recognition fidelity. Beyond efficiency, Youtu-Parsing exhibits superior robustness in parsing heterogeneous document elements, ranging from intricate formulas and tables to administrative seals, even under adverse conditions such as handwritten annotations or rare characters. Extensive evaluations on authoritative benchmarks, including OmniDocBench and olmOCR-bench, demonstrate that Youtu-Parsing achieves state-of-the-art performance, surpassing both general-purpose large multimodal models and specialized parsing systems. These results establish Youtu-Parsing as a robust and scalable foundation for large-scale information extraction and advanced downstream knowledge management.

\clearpage
\section*{Contributions and Acknowledgments}

We would like to express our sincere gratitude to all contributors, including those not listed in the paper, for their invaluable support and efforts. \textbf{The contributors within each group are listed in no particular order.}

\textbf{Core Contributors}\\
Haoyu Cao \quad
Kun Yin \quad
Yunfei Wu \quad
Bing Liu \quad
Zhongpeng Cai \quad
Xiaotian Li \quad
Huang Chen \quad
Xin Li \quad
Yinsong Liu \quad
Deqiang Jiang \quad
Xing Sun$^\dagger$ \quad
Yunsheng Wu \quad
\noindent\let\thefootnote\relax\footnotetext{$^\dagger$Corresponding author: winfredsun@tencent.com}
\vspace{1em}

\textbf{Contributors}\\
Qianyu Li \quad
Antai Guo \quad
Yanzhen Liao \quad
Yanqiu Qu \quad
Haodong Lin \quad
Chengxu He \quad
Shuangyin Liu \quad

\setcitestyle{square,numbers,comma}
\bibliography{youtu_bib}

\clearpage

\appendix

\section*{Appendix}

\section{Related Work}

\subsection{Traditional pipeline-based methods}

Conventional document parsing systems primarily adopt a pipeline architecture that decomposes the overall task into a sequence of discrete, modular components. These components typically include layout detection \citep{zhao2024doclayout,sun2025pp}, text and formula recognition \citep{zhou2022end}, table structure extraction \citep{siddiqui2019deeptabstr, lin2022tsrformer}, and reading order inference \citep{wang2021layoutreader}. Each module is usually implemented via specialized algorithms or models optimized for its specific sub-task, facilitating targeted improvements and flexible integration within the overall system. For instance, systems like Marker \citep{Marker} utilize dedicated OCR engines, layout analyzers, and table recognition modules to handle diverse document formats, often complemented by Large Language Models (LLMs) to enhance capabilities such as cross-page table merging and inline formula parsing. Similarly, frameworks like MinerU \citep{wang2024mineru} and PP-StructureV3 \citep{cui2025paddleocr} orchestrate sequential processes involving layout analysis, content region segmentation, recognition, and reading order prediction to generate structured representations of documents. The modular design allows for isolated development and maintenance, enabling researchers and practitioners to improve individual components independently.

\subsection{Generalized Vision-Language Models}
Recent developments in vision-language modeling have significantly advanced the field of comprehensive document understanding, providing promising alternatives to traditional OCR pipelines. Modern models, such as Gemini2.5 Pro \citep{comanici2025gemini}, GPT-4o \citep{hurst2024gpt}, InternVL series \citep{chen2024internvl, chen2024far, wang2024mpo, chen2024expanding, wang2025internvl3_5} and QwenVL series \citep{Qwen-VL, Qwen2VL, Qwen2.5-VL, qwen3technicalreport}, demonstrate strong proficiency in extracting both textual and symbolic information directly from document images. These models leverage large-scale pretraining on diverse multimodal datasets, enabling them to bridge the gap between generic visual and language understanding frameworks and domain-specific OCR tasks. Their versatility extends across various document types, languages—including multilingual and handwritten scripts—and challenging layouts, such as complex formatting or low-quality images.

\subsection{Task-specific visual-language models}
In response to the limitations of both traditional pipelines and large generalized models, recent research has explored task-specific and hybrid approaches to document understanding \citep{poznanski2025olmocr, li2025dots, niu2025mineru2, cui2025paddleocrvl}, aiming for a balance between accuracy, efficiency, and modularity. These approaches often employ end-to-end vision-language architectures designed to handle multiple document elements simultaneously, including text, tables, figures, and layout information. Early efforts in this domain established the feasibility of unified frameworks capable of processing heterogeneous document modalities within a single model, thus facilitating joint representation learning and inter-element reasoning. Building upon this foundation, more recent models leverage high-capacity vision encoders trained on large-scale, domain-relevant datasets to improve extraction fidelity and contextual understanding, enabling more accurate recognition and structuring.

\clearpage

\section{Qualitative examples}

\subsection{Text Recognition}

\subsubsection{Printed Text Recognition}

\begin{figure}[h]
\begin{center}
\includegraphics[width=1.0\textwidth]{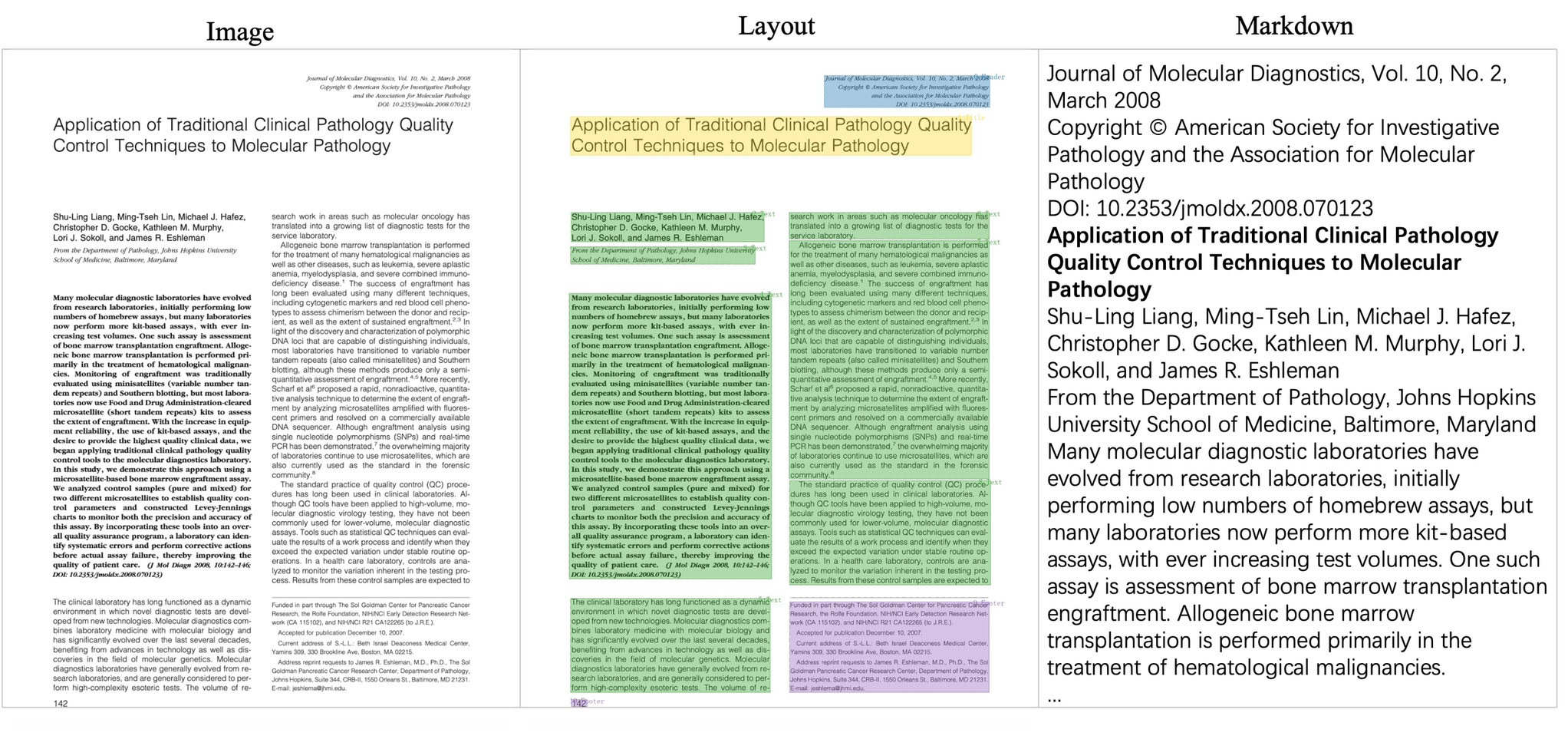}
\end{center}
\end{figure}

\subsubsection{Handwriting Text Recognition}

\begin{figure}[h]
\begin{center}
\includegraphics[width=1.0\textwidth]{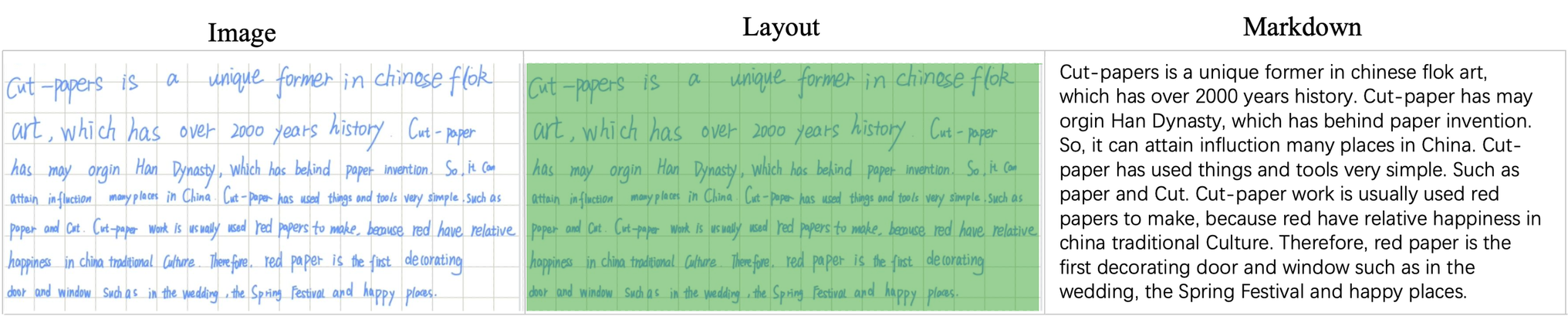}
\end{center}
\end{figure}

\begin{figure}[h]
\begin{center}
\includegraphics[width=1.0\textwidth]{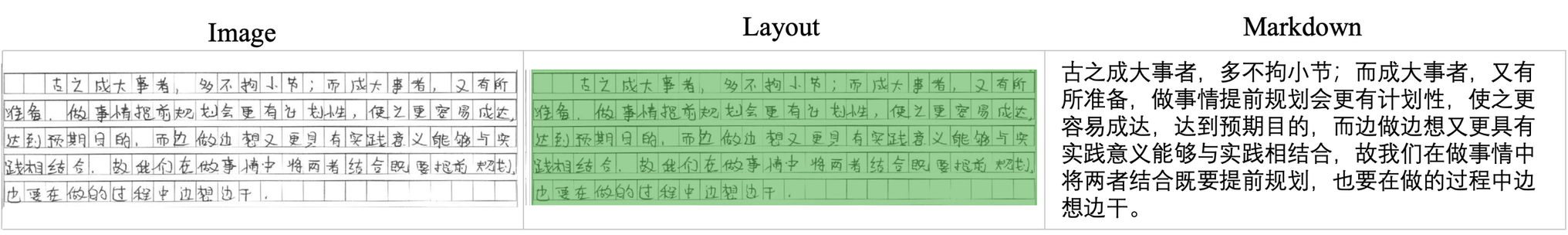}
\end{center}
\end{figure}

\clearpage
\subsubsection{Multilingual Text Recognition}

\begin{figure}[h]
\begin{center}
\includegraphics[width=1.0\textwidth]{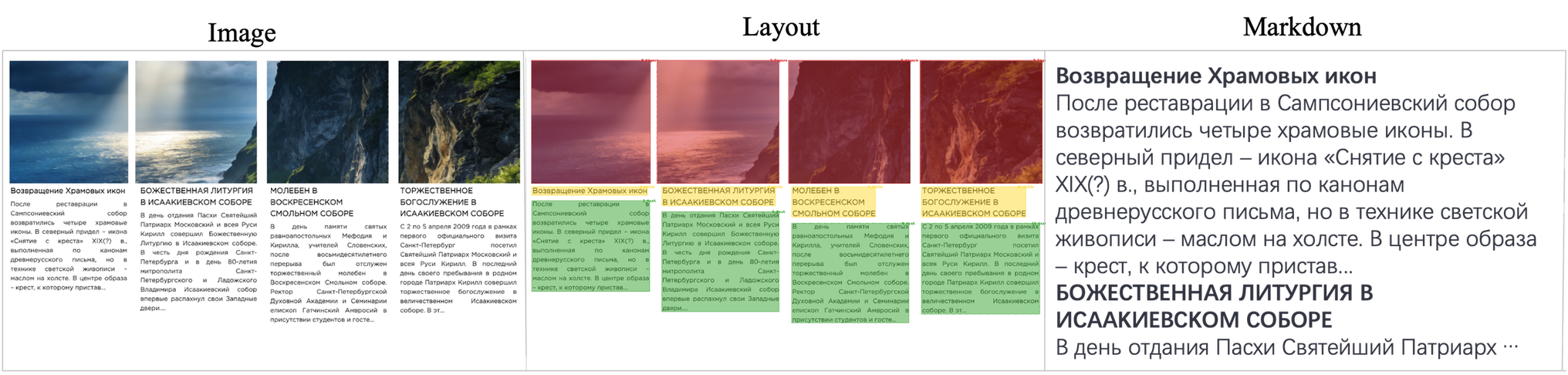}
\end{center}
\end{figure}

\subsubsection{Rare Char Recognition}

\begin{figure}[h]
\begin{center}
\includegraphics[width=1.0\textwidth]{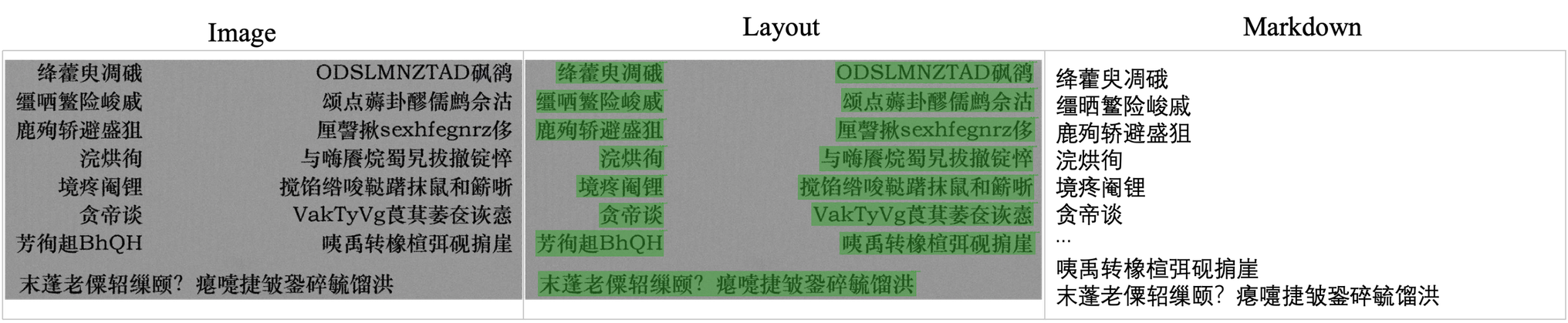}
\end{center}
\end{figure}

\subsubsection{Seal Recognition}
\begin{figure}[h]
\begin{center}
\includegraphics[width=1.0\textwidth]{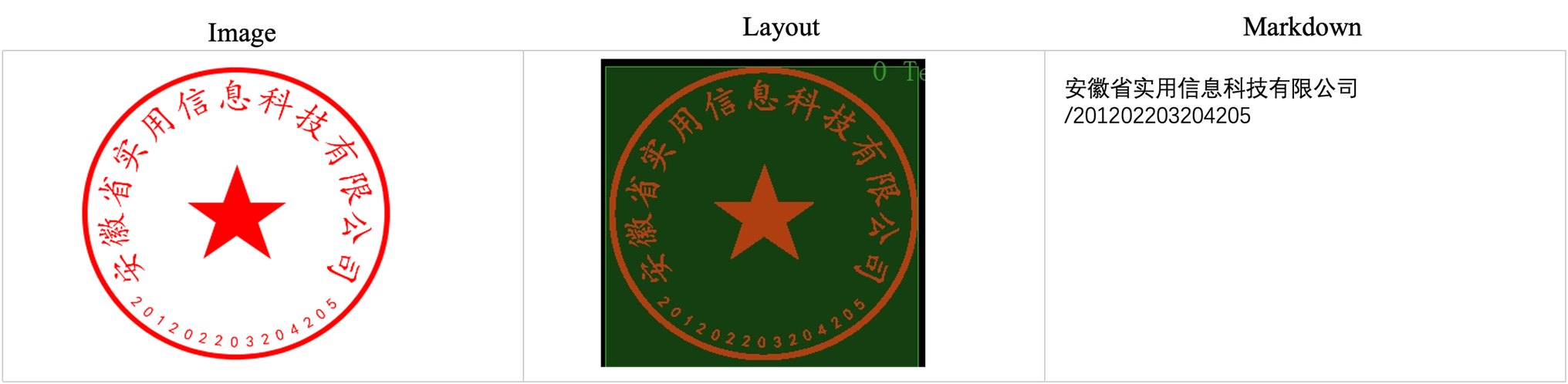}
\end{center}
\end{figure}

\clearpage
\subsubsection{Art Font Text Recognition}

\begin{figure}[h]
\begin{center}
\includegraphics[width=1.0\textwidth]{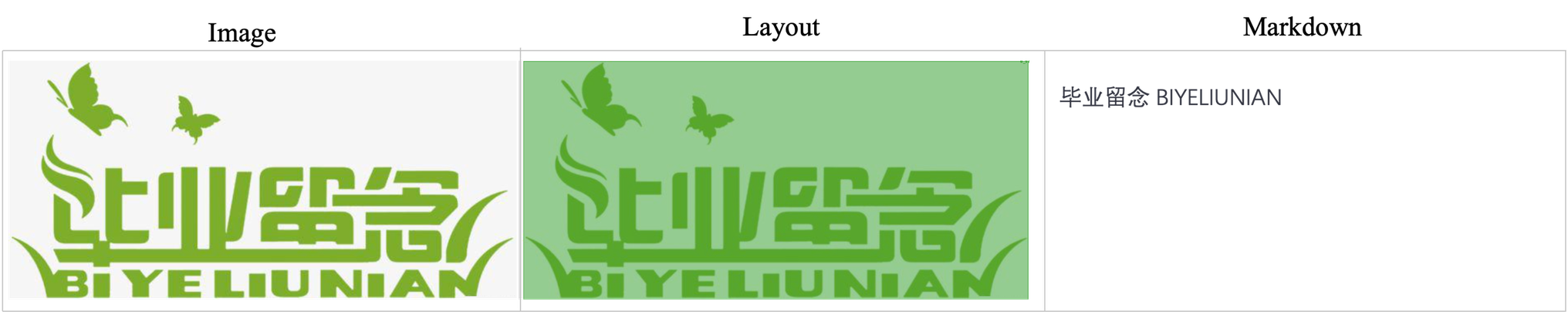}
\end{center}
\end{figure}

\begin{figure}[h]
\begin{center}
\includegraphics[width=1.0\textwidth]{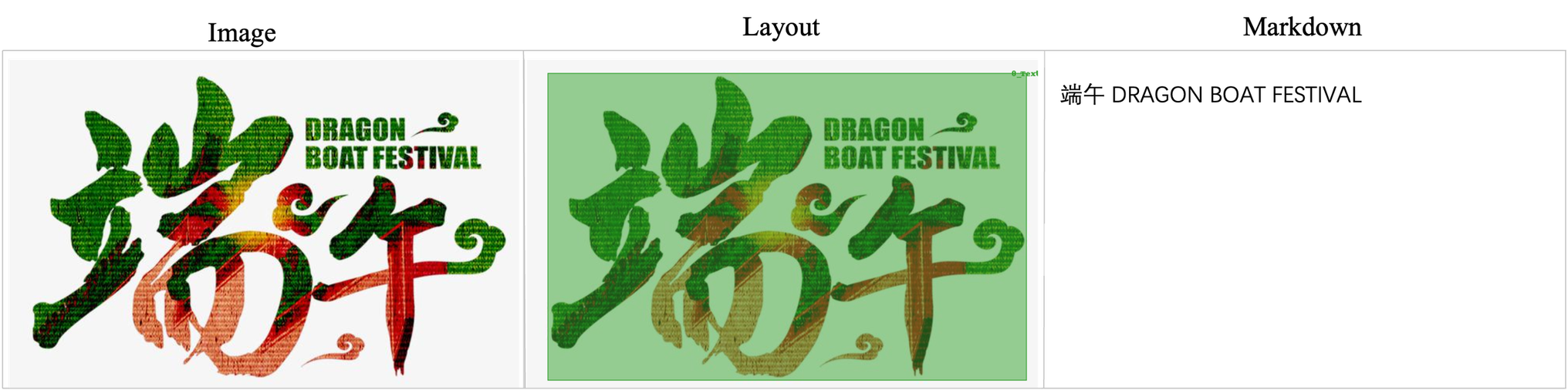}
\end{center}
\end{figure}

\subsubsection{Vertical Text Recognition}

\begin{figure}[h]
\begin{center}
\includegraphics[width=1.0\textwidth]{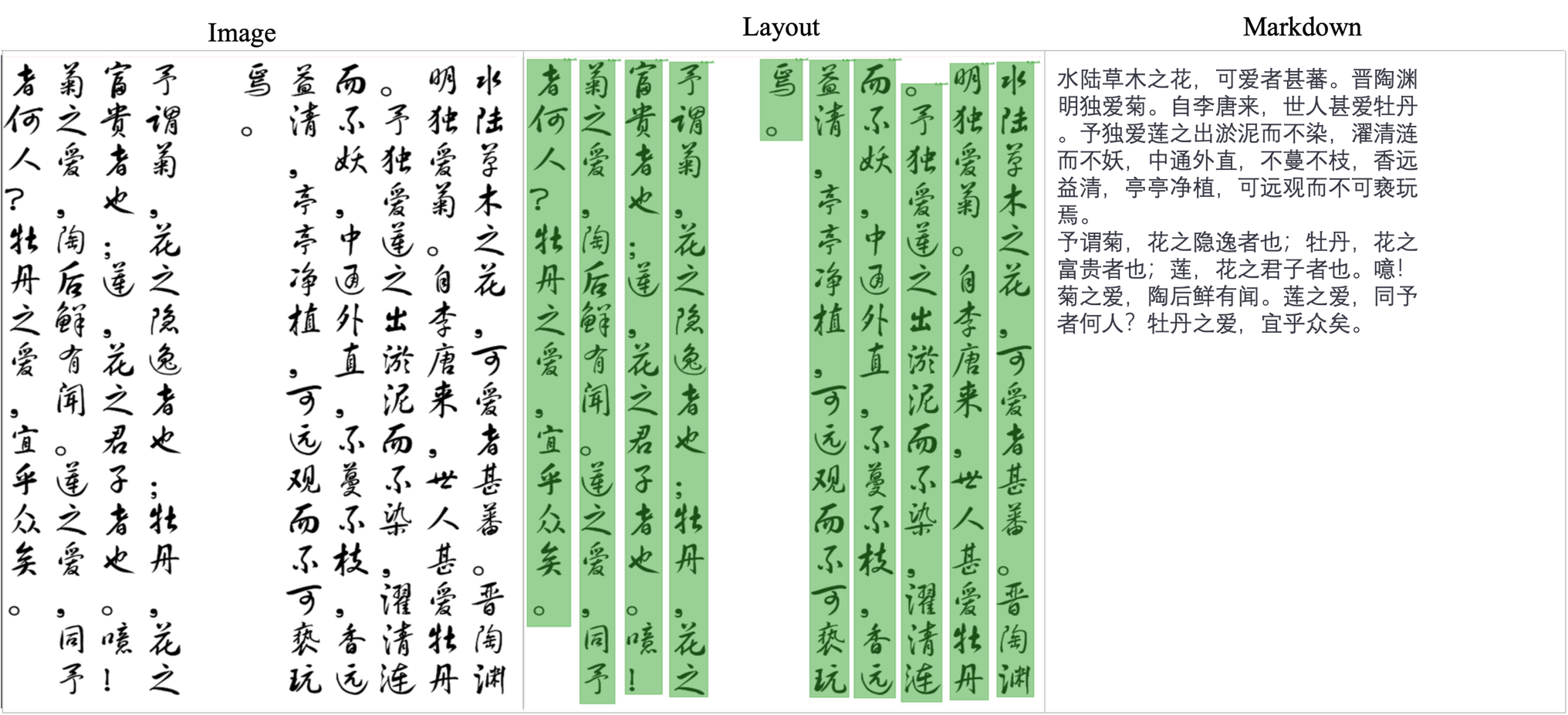}
\end{center}
\end{figure}

\clearpage
\subsection{Table Recognition}

\subsubsection{Wired Table Recognition}

\begin{figure}[h]
\begin{center}
\includegraphics[width=1.0\textwidth]{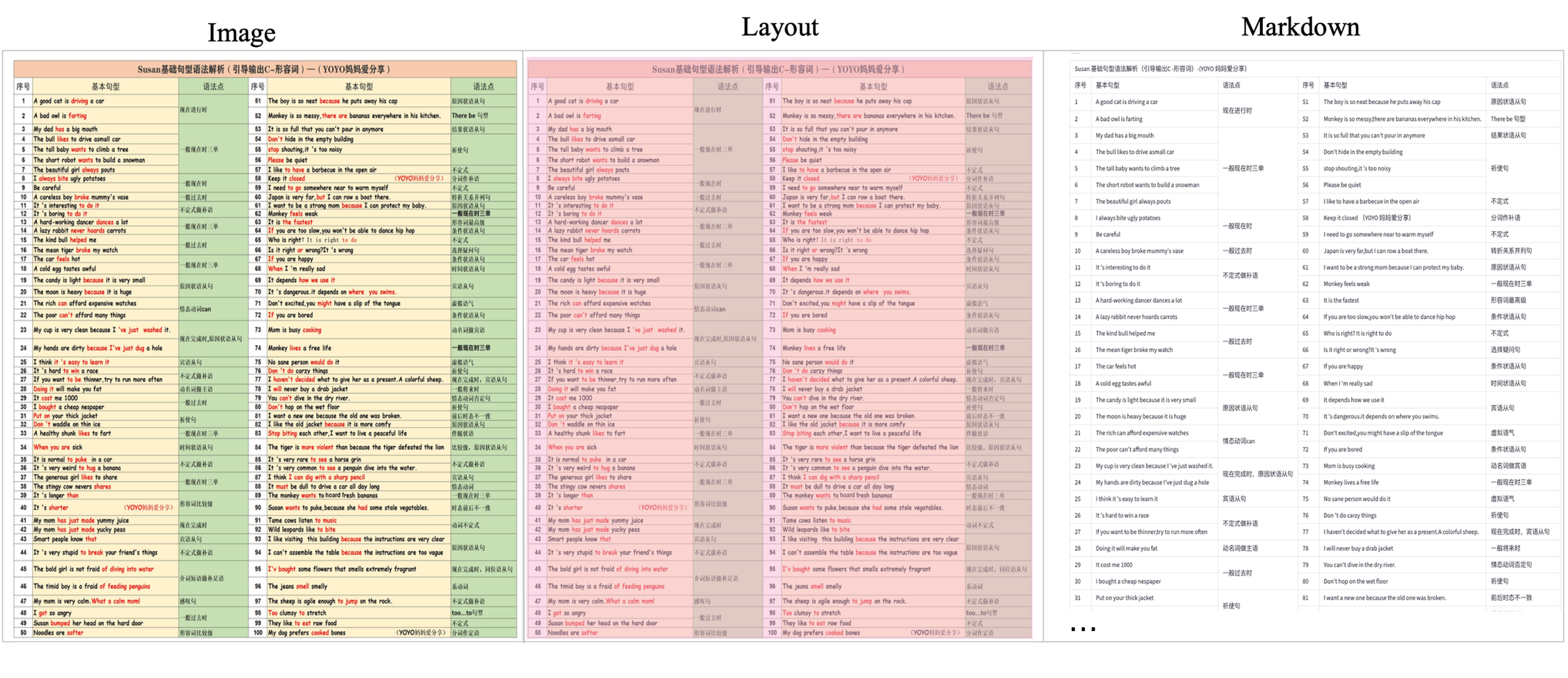}
\end{center}
\end{figure}

\subsubsection{Wireless Table Recognition}

\begin{figure}[h]
\begin{center}
\includegraphics[width=1.0\textwidth]{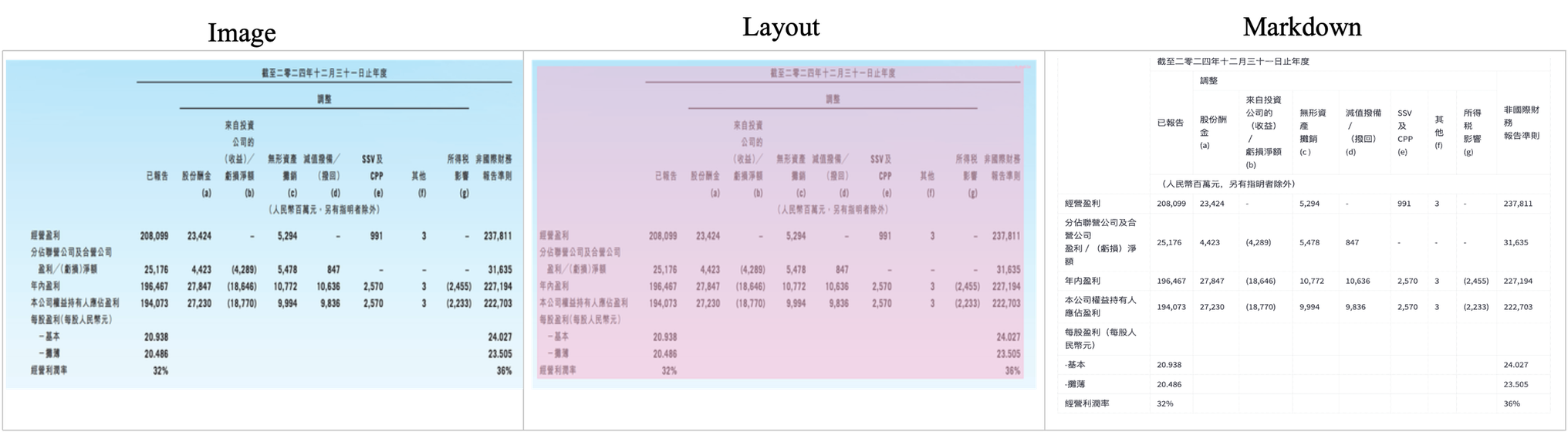}
\end{center}
\end{figure}

\clearpage
\subsection{Formula Recognition}

\subsubsection{Multi-line Formula}
\begin{figure}[h]
\begin{center}
\includegraphics[width=1.0\textwidth]{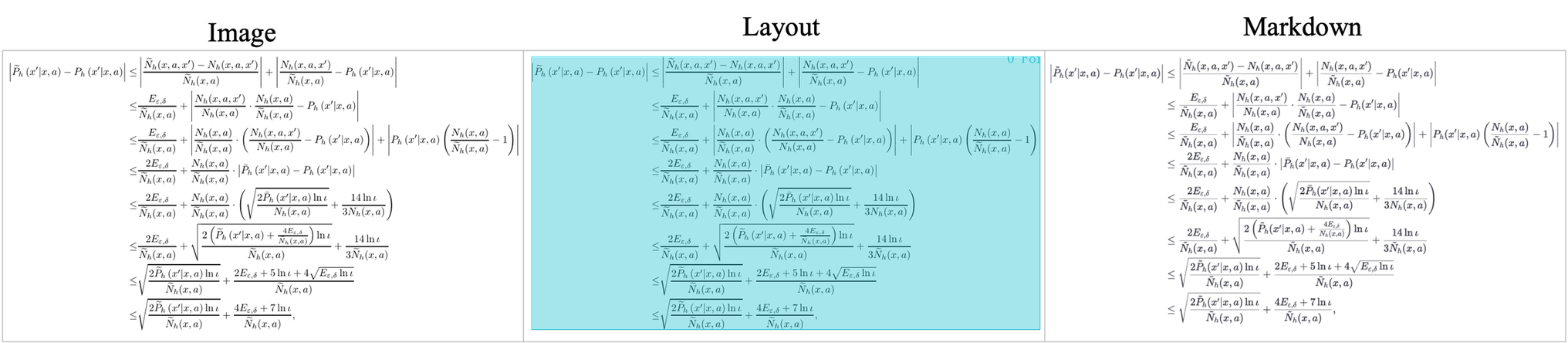}
\end{center}
\end{figure}

\subsubsection{Matrices}
\begin{figure}[h]
\begin{center}
\includegraphics[width=1.0\textwidth]{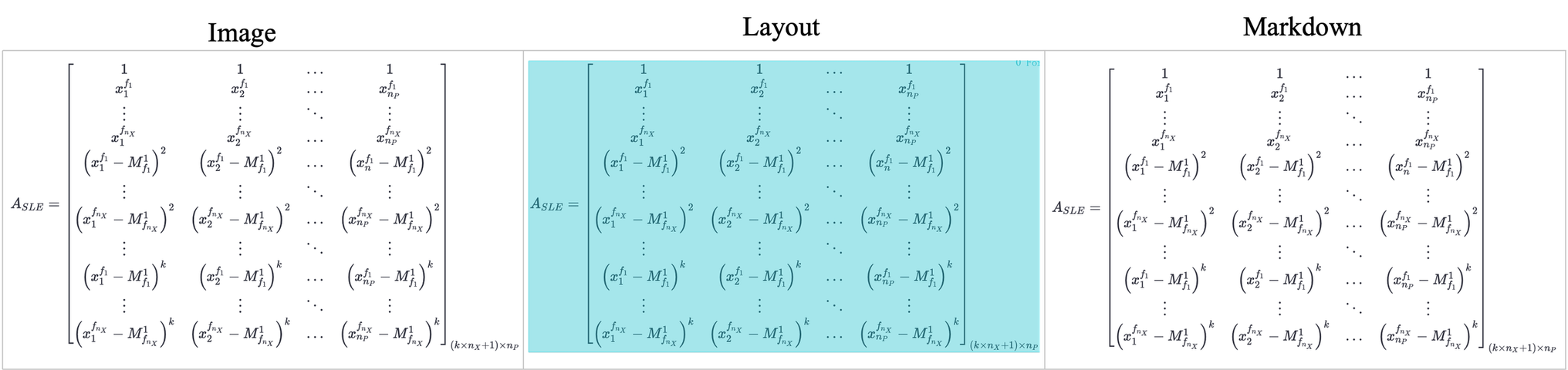}
\end{center}
\end{figure}

\subsubsection{Handwritten Formula}
\begin{figure}[h]
\begin{center}
\includegraphics[width=1.0\textwidth]{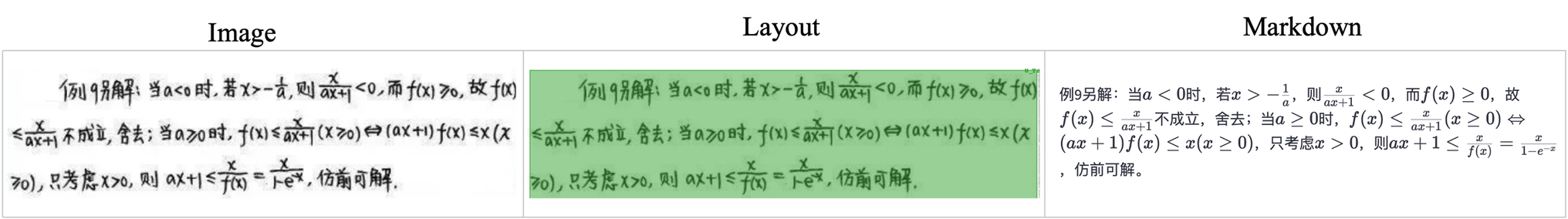}
\end{center}
\end{figure}

\clearpage
\subsection{Hierarchical Structure Analysis}
\subsubsection{Group Relation}
\begin{figure}[h]
\begin{center}
\includegraphics[width=1.0\textwidth]{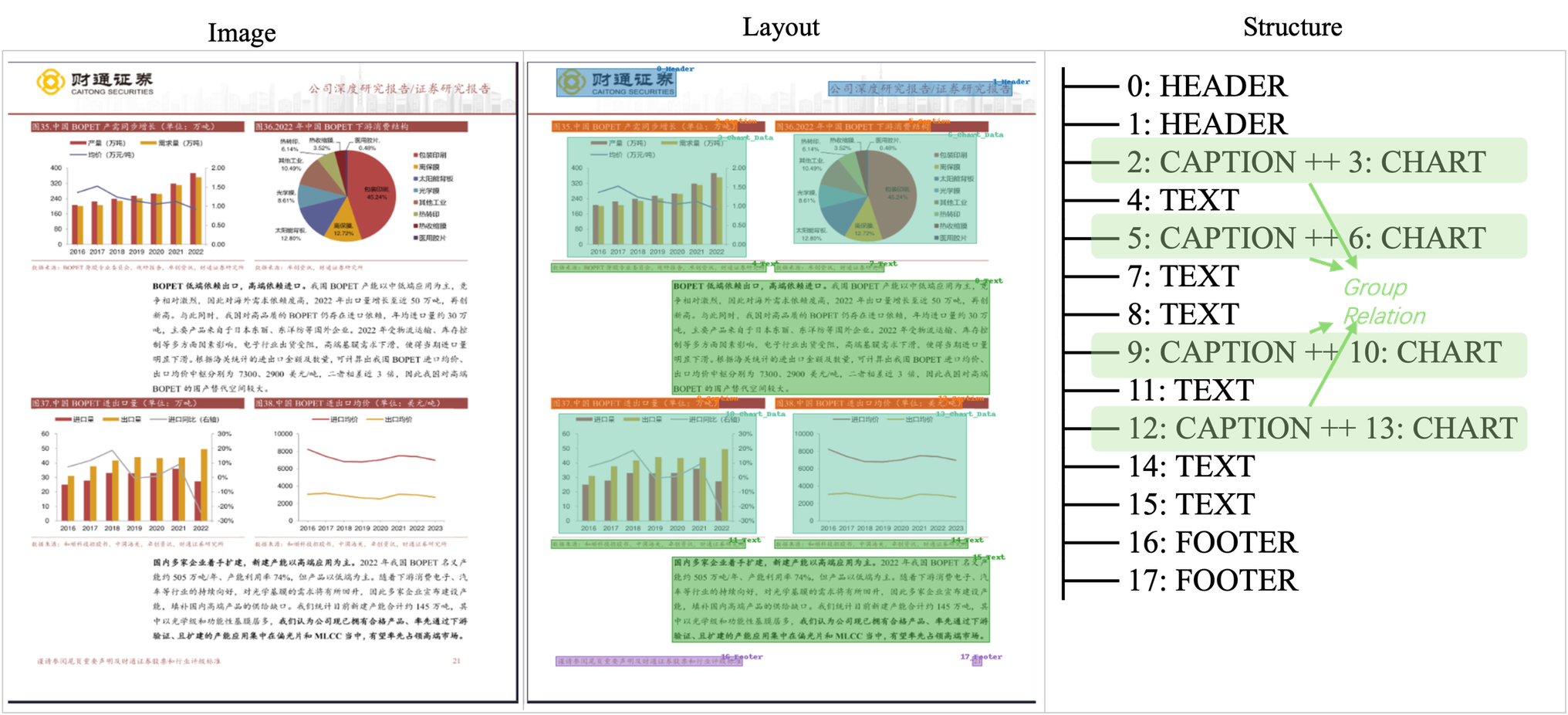}
\end{center}
\end{figure}

\subsubsection{Parent-Child Relation}
\begin{figure}[h]
\begin{center}
\includegraphics[width=1.0\textwidth]{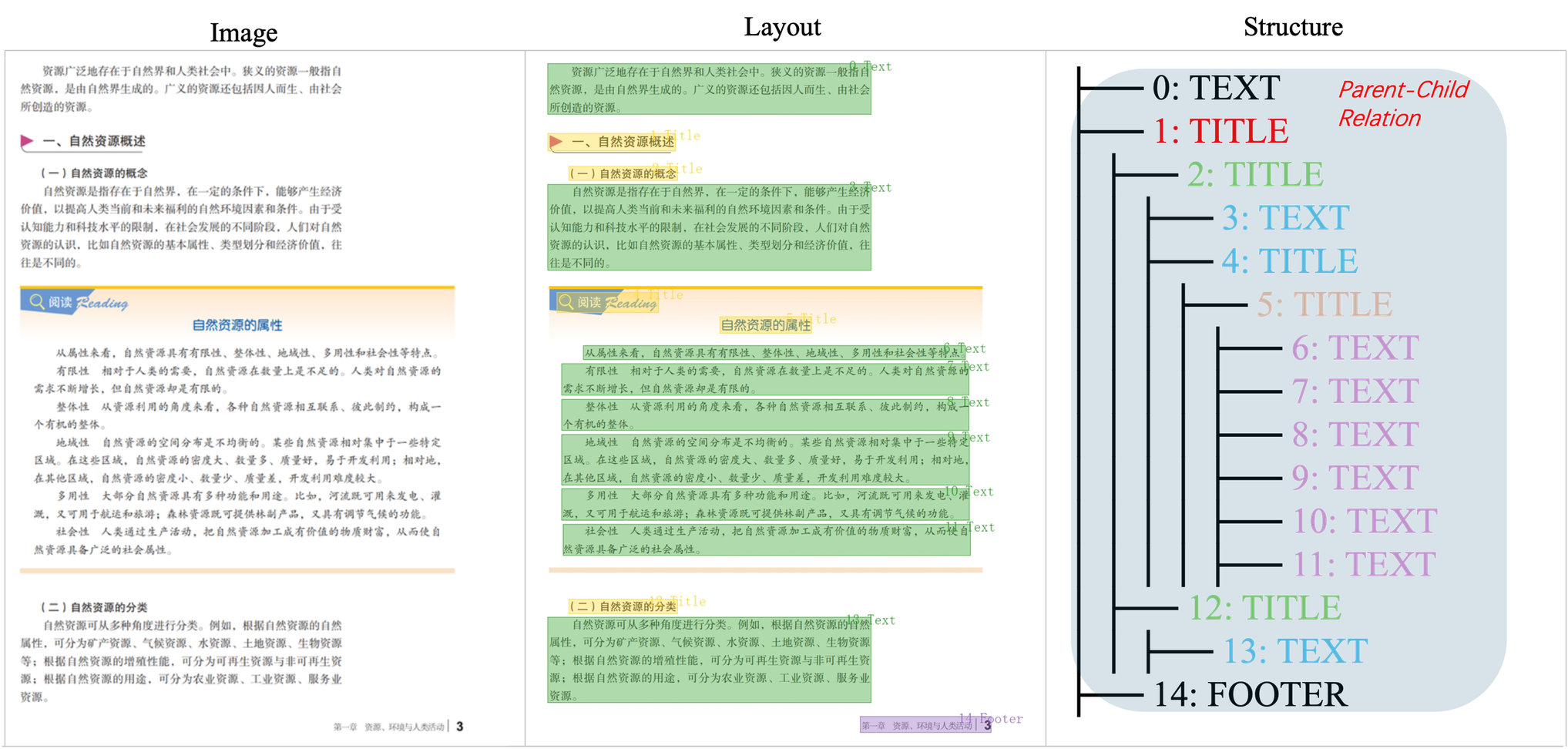}
\end{center}
\end{figure}

\clearpage
\subsection{Chart Recognition}
\begin{figure}[H]
\begin{center}
\includegraphics[width=1.0\textwidth]{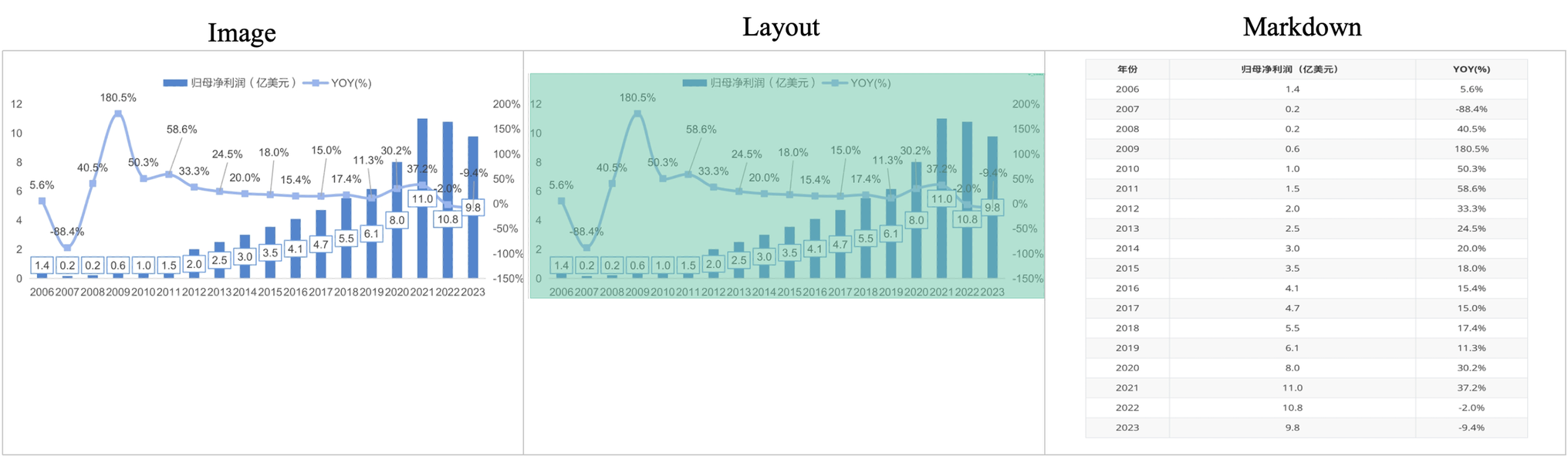}
\end{center}
\end{figure}

\begin{figure}[H]
\begin{center}
\includegraphics[width=1.0\textwidth]{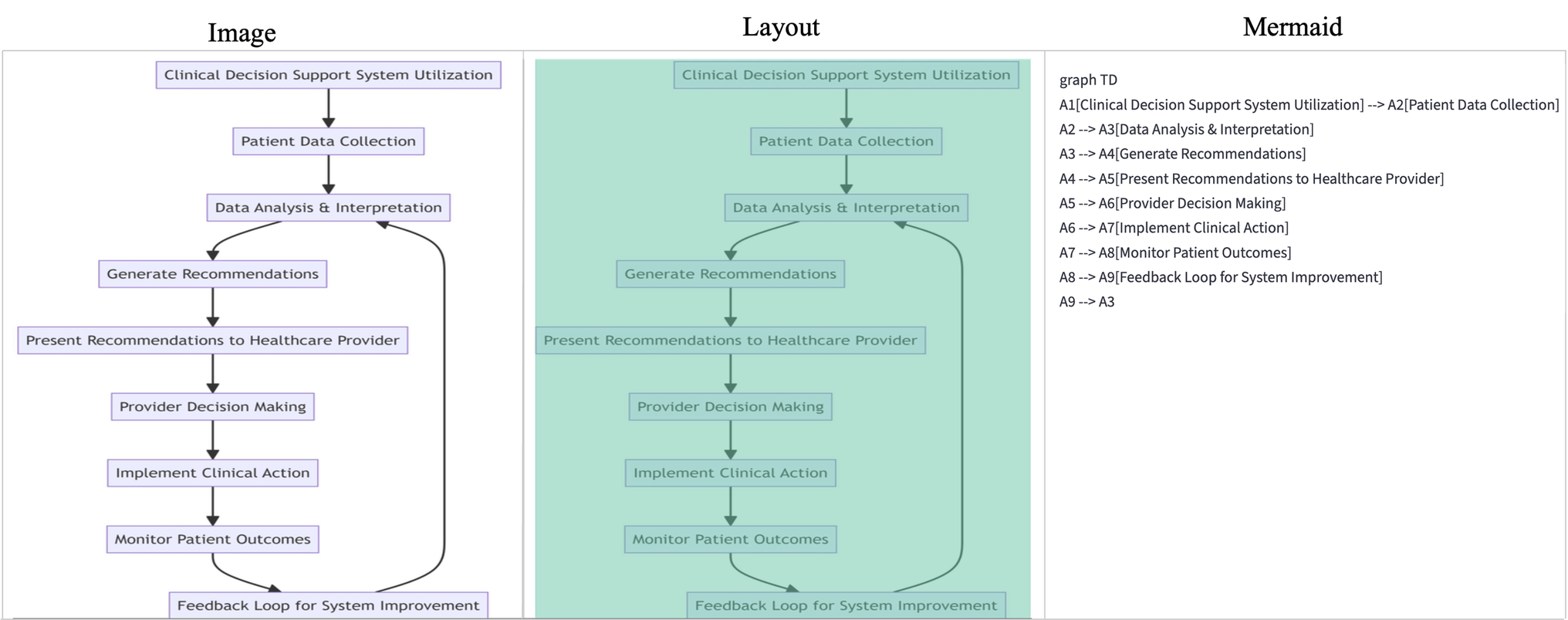}
\end{center}
\end{figure}

\begin{figure}[H]
\begin{center}
\includegraphics[width=1.0\textwidth]{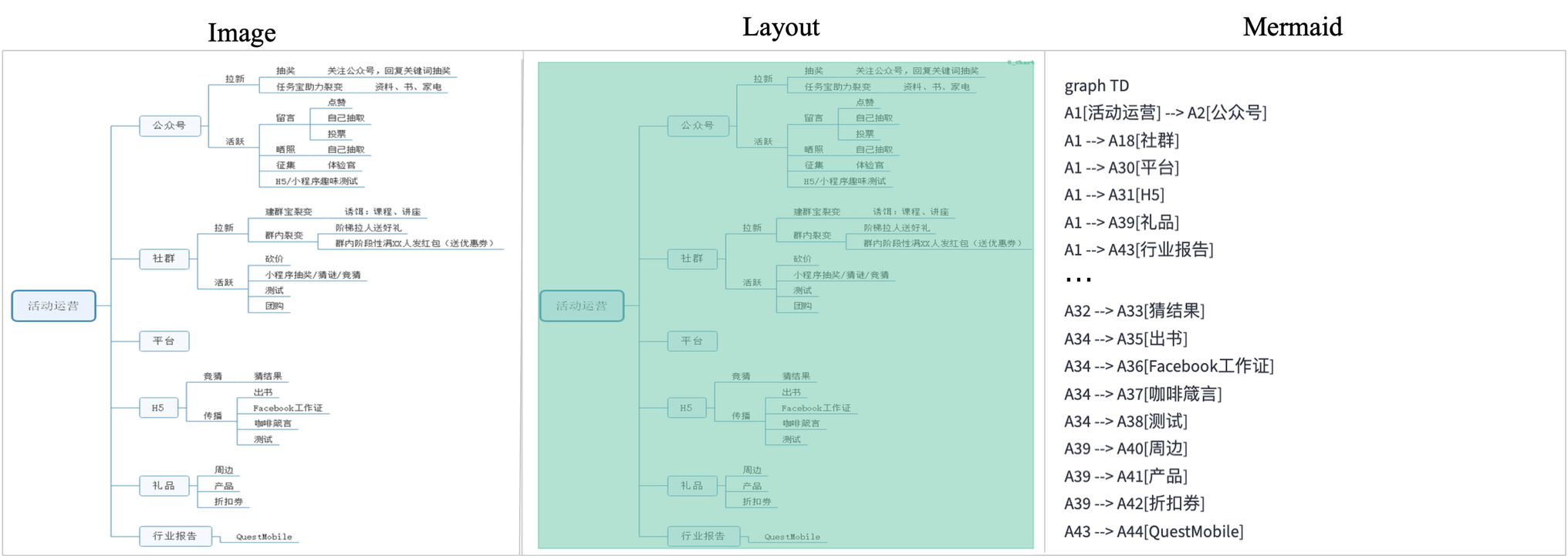}
\end{center}
\end{figure}

\clearpage

\section{Prompt Details}
In this appendix, we provide a comprehensive description of the prompts employed throughout the two-stage inference pipeline of Youtu-Parsing. These prompts are carefully designed to guide the model in performing various document understanding tasks, including layout detection, hierarchy recognition, text recognition, chart parsing, figure text recognition, and OCR Task.

\begin{table}[htbp]
\centering
\caption{Prompts used in Youtu-Parsing for different document understanding tasks.}
\label{tab:prompts}
\renewcommand{\arraystretch}{1.4}
\small
\begin{tabular}{l>{\centering\arraybackslash}c>{\raggedright\arraybackslash}p{9.5cm}}
\toprule
\textbf{Task} & \textbf{Lang} & \textbf{Prompt} \\
\midrule
\multirow{2}{*}[0.5em]{\textbf{Layout Detection}}
& CN & 分析输入文档的版面结构，检测所有结构化元素并进行语义分类。使用``\textbackslash n''区分不同区域的边界。 \\[2ex]
& EN & Analyze the layout structure of the input document, detect all structural elements and classify them semantically. Use \textbackslash n to delimit different regions. \\[2ex]
\midrule
\multirow{2}{*}[0.5em]{\textbf{Hierarchy Recognition}} 
& CN & 识别输入框间的层级关系，输出层级结构。 \\[2ex]
& EN & Identify hierarchical relationships among input fields and output the hierarchy structure. \\[2ex]
\midrule
\multirow{2}{*}[0.5em]{\textbf{Text Recognition}}
& CN & 根据给定的输入框坐标与版面类型，识别并提取该区域内的内容。其中，公式以LaTeX格式输出，表格以OTSL格式输出：\texttt{<x\_\{\}><y\_\{\}><x\_\{\}><y\_\{\}><LAYOUT\_\{\}>} \\[2ex]
& EN & Based on the given input field coordinates and layout type, identify and extract the content within the specified region. Formulas shall be represented in LaTeX notation, and tables shall be structured in OTSL format: \texttt{<x\_\{\}><y\_\{\}><x\_\{\}><y\_\{\}><LAYOUT\_\{\}>} \\[2ex]
\midrule
\multirow{2}{*}[0.5em]{\textbf{Chart Parse}} 
& CN & 将图中的逻辑类图表用Mermaid格式输出，数据类图表用Markdown格式输出。 \\[2ex]
& EN & Convert the logic charts in the figure to Mermaid format and the data charts to Markdown format. \\[2ex]
\midrule
\multirow{2}{*}[0.5em]{\textbf{Figure Text Recognition}} 
& CN & 提取插图的所有文本元素，并以结构化形式输出识别结果。 \\[2ex]
& EN & Extract all textual elements from the given figure and present the recognition results in a structured manner. \\[2ex]
\midrule
\multirow{2}{*}[0.5em]{\textbf{OCR Task}}
& CN & 识别图像中的全部文本内容，保持原始文档的层级结构与排版逻辑。 \\[2ex]
& EN & Recognize all textual content in the image and output the results in a structured format. \\[2ex]
\bottomrule
\end{tabular}
\end{table}

Note: The prompt for each task is tailored to the specific requirements of the task, and may vary slightly between different languages. Additionally, the layout types corresponding to text recognition can be \texttt{TEXT}, \texttt{TITLE}, \texttt{TABLE}, \texttt{FIGURE}, \texttt{CHART\_DATA}, \texttt{CHART\_LOGIC}, \texttt{SEAL}, \texttt{HEADER\_FIGURE}, \texttt{FOOTER\_FIGURE} (where \texttt{FORMULA}, \texttt{HEADER}, \texttt{FOOTER}, \texttt{CAPTION}, and \texttt{CODE} are all normalized to the \texttt{TEXT} category).

\end{CJK}
\end{document}